\documentclass[twoside]{article}
\usepackage{amsmath,amssymb,amsfonts}
\usepackage[accepted]{aistats2023}
\usepackage{tikz}
\usepackage{mathtools}
\usepackage{algorithmic}
\usepackage{graphicx}
\usepackage{textcomp}
\usepackage{xcolor}
\usepackage{adjustbox}
\usepackage{caption}
\usepackage{subcaption}
\usepackage{dsfont}
\usepackage{array}
\usepackage{units}
\usepackage{graphicx,multirow}
\newcolumntype{C}{>{\centering\arraybackslash}p{1em}}
\usepackage[ruled,linesnumbered]{algorithm2e}
\usepackage{graphicx}
\usepackage{natbib}

\usepackage{xr-hyper}
\usepackage[colorlinks=true, allcolors=blue]{hyperref}
\begin{document}

\twocolumn[

\aistatstitle{EGG-GAE: scalable graph neural networks for tabular data imputation}

\aistatsauthor{Lev Telyatnikov \\
              Sapienza University of Rome \\
              \And 
              Simone Scardapane \\
              Sapienza University of Rome }

\aistatsaddress{ lev.telyatnikov@uniroma1.it  \And simone.scardapane@uniroma1.it } ]

\begin{abstract}
  Missing data imputation (MDI) is crucial when dealing with tabular datasets across various domains. Autoencoders can be trained to reconstruct missing values, and \textit{graph} autoencoders (GAE) can additionally consider similar patterns in the dataset when imputing new values for a given instance. However, previously proposed GAEs suffer from scalability issues, requiring the user to define a similarity metric among patterns to build the graph connectivity beforehand. In this paper, we leverage recent progress in latent graph imputation to propose a novel EdGe Generation Graph AutoEncoder (EGG-GAE) for missing data imputation that overcomes these two drawbacks. EGG-GAE works on randomly sampled mini-batches of the input data (hence scaling to larger datasets), and it automatically infers the best connectivity across the mini-batch for each architecture layer. We also experiment with several extensions, including an ensemble strategy for inference and the inclusion of what we call \textit{prototype nodes}, obtaining significant improvements, both in terms of imputation error and final downstream accuracy, across multiple benchmarks and baselines.
\end{abstract}

\begin{figure*}[ht]
\centering
\includegraphics[width=0.9\textwidth]{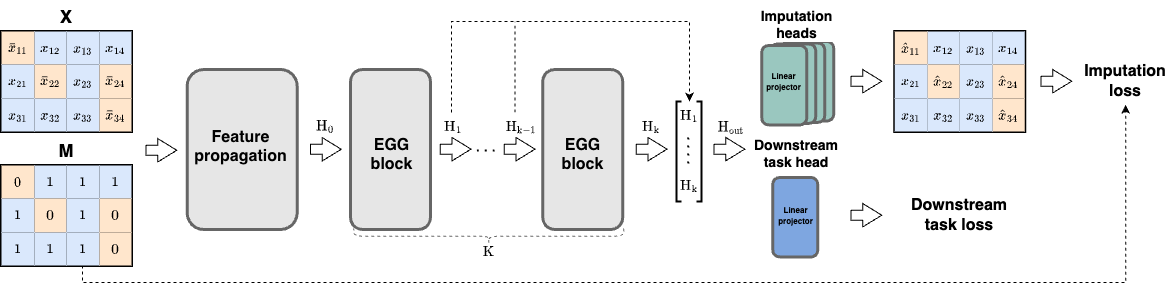}
\caption{Pipeline:  initially, the preprocessed batch $\mathrm{X}$ is encoded with a feature propagation block (Eq. \eqref{FeaturePropagationEq}), resulting in the initial representations $\mathrm{H}_{\mathrm{0}} $. These representations are then updated with a sequence of EGG-GAE blocks obtaining ${\mathrm{H}_{\mathrm{k}}}$, $\mathrm{k} \in \{\mathrm{1}, \dots, \mathrm{K}\}$. The concatenation of $\{\mathrm{H}_{\mathrm{k}}\}_{\mathrm{k}=\mathrm{1}}^{\mathrm{K}}$ gives the final representation $\mathrm{H}_{\mathrm{out}}$. It is used as the input for the subsequent heads: numerical imputation, categorical imputation, and downstream task one.}
\label{fig:model_pipline}
\end{figure*}

\section{INTRODUCTION}

Missing data imputation is a ubiquitous issue that arises in a variety of domains. Most supervised deep learning methods require complete datasets, but in the real world, datasets often suffer from incompleteness due to access problems or mistakes in data collection \citep{yoon2016discovery,yoon2018personalized,kreindler2006effects,van2018flexible}. Numerous fields thus require missing data imputation (MDI) methods to reconstruct a complete dataset, including biostatistics \citep{mackinnon2010use}, epidemiology \citep{sterne2009multiple}, and irregular time-series analysis \citep{kreindler2006effects}. 

Classically, the underlying mechanism giving rise to missing data is categorized into three types \citep{rubin1976inference}.  (1) If the probability of being missing is completely independent of the data, then the data are said to be missing at random (MCAR). (2) Data is said to be missing at random (MAR) if the probability of being missing is the same only within the data-defined groups. (3) If the probability of missing data depends on both observed and unobserved variables, then such data are missing not at random (MNAR). 

Predictive approaches to MDI \citep{bertsimas2017predictive} can be categorized in two families: (i) building a global model for data imputation, or (ii) inferring the missing components employing similar data points to the one having missing values. The second class typically uses advanced k-NN strategies \citep{acuna2004treatment}. The first class includes simple statistics from the entire dataset (e.g., medians), linear models \citep{lakshminarayan1996imputation}, support vector machines \citep{wang2006missing} or, more recently, deep neural architectures \citep{smieja2018processing,yoon2018gain, nazabal2020handling,spinelli2020missing}. 

Recently, it was noted that the unification of the two paradigms (i.e., inferring similar data points for each imputation and building global models from the overall dataset) can be beneficial for MDI. This can be done by exploiting Graph Neural Networks (GNNs) \citep{narang2013signal,chen2020learning, jiang2020incomplete,rossi2021unreasonable}, a novel class of neural networks that can process graph-based data in a differentiable fashion. In particular, the graph imputer neural network (GINN) \citep{spinelli2020missing} explored the assumption of endowing tabular data with a graph topology based on a pre-computed similarity between points in the feature space, and then exploiting a GNNs to tackle the MDI problem. However, the proposed GINN model had two limitations. First, the method is tested only on the entire dataset (i.e., no mini-batching is performed, which is challenging with GNNs models), and scaling it up to large datasets is unfeasible due to the quadratic cost of computing the similarity matrix. Second, graph connectivity must be computed beforehand in the feature space employing only those values observed by both data points, requiring the definition of a suitable distance metric and a customized procedure to sparsify the graph.

In general, building a connectivity beforehand as done in GINN was necessary, since the underlying assumption of most GNNs is that the graph topology is given and fixed; thus, convolution-like operations typically amount to modifying the node-wise features by averaging information from the neighbours. However, the hand-crafted connectivity might be sub-optimal and requires a number of hyper-parameters to be defined and manually optimized. Instead, automatically learning the latent graph structure can overcome the limitations of these methods by inferring the underlying graph relationships. Such latent graphs can capture the actual topology of structured data through the downstream tasks, which can be seen as a task-related topology, thus conveying model interpretability. Recently, latent graph imputation has become an important research topic in the GNNs literature, as described in Section \ref{subsec:learning_graph_structure}.

Based on these considerations, the main contributions of this paper are as follows. (i) We introduce an end-to-end trainable network architecture to learn the optimal underlying latent graph for tabular data using a novel EdGe Generation Graph AutoEncoder (EGG-GAE) network module. The EGG-GAE module sampling scheme is optimized with respect to downstream task metrics utilizing a straight-through estimator \citep{jang2016categorical} in the backward pass to ensure its differentiability. (ii) We demonstrate that employing the latent graph predictions for the missing data imputation (MDI) problem induces consistent improvement across a number of datasets in a large experimental evaluation, demonstrating significant improvements over baselines and reaching state-of-the-art performance. (iii) We propose the concept of tabular graph mini-batching along with an ensembling technique to resolve the MDI scalability issues \citep{miao2022experimental}. (iv) We propose a novel concept of learnable \emph{prototype nodes} which encodes a learnable data representation in the form of an additional set of nodes added to the imputed graph, to provide each data point in the mini-batch with a reliable neighbourhood.

\section{RELATED WORKS}
\label{sec:related_works}

Before describing our proposed solution for tabular MDI, we provide a brief overview of related works among three lines: MDI methods for tabular data (Section \ref{subsec:tabular_mdi}), latent graph imputation for GNNs (Section \ref{subsec:learning_graph_structure}), and methods to employ GNNs on graphs with missing data (Section \ref{subsec:gnns_node_imputation}).

\subsection{Tabular missing data imputation}
\label{subsec:tabular_mdi}

MDI algorithms can be categorized depending on whether they are discriminative or generative, univariate or multivariate, and on whether they provide one or multiple imputations for each missing data point. In this work we present a generative model which performs multivariate imputations and can provide multiple imputations for each missing datum.

The imputation strategies can be divided into three categories: 1) statistical; 2) machine learning (ML), and 3) deep learning (DL) based. Statistical methods exploit the observed data to obtain mean, median, or mode estimation of missing data points \citep{farhangfar2007novel}. Among traditional machine learning imputation approaches, multiple imputation using chained equations (MICE) \citep{van2011mice} is considered one of the most flexible and powerful. MICE iteratively imputes each dataset variable while keeping the others constant, selecting one or more observations from a predictive distribution on that variable. Although MICE has performed well in some cases, its underlying assumptions may result in biased forecasts and lower accuracy \citep{azur2011multiple}. ML algorithms include k-nearest neighbours (KNN) \citep{acuna2004treatment}, decision trees \cite{lakshminarayan1996imputation}, support vector techniques \citep{wang2006missing} and several others. In practic

e, these approaches have mixed results compared to more straightforward techniques such as mean imputation \citep{bertsimas2017predictive}. KNN is generally limited to weighted averaging among similar feature vectors, whereas other algorithms are required to build a global dataset model for imputation. DL models include deep denoising autoencoders \citep{gondara2018mida}, recurrent neural networks \citep{bengio1995recurrent}, and generative models \citep{yoon2018gain, nazabal2020handling}. Multilayer nonlinear computation allows these methods to capture more complex correlations in data, however, they still require building a global model from the dataset while ignoring potentially significant contributions from similar points. GINN \citep{spinelli2020missing} addressed the problem of leveraging both the global aspect of the dataset and local dependencies between different data points utilizing GNNs. GINN requires the calculation of a pre-defined similarity matrix on the entire dataset in feature space, which is unfeasible for most real-world databases.

\subsection{Latent graph learning}
\label{subsec:learning_graph_structure}

GNNs exploit the general idea of localized message-passing, e.g., using graph convolutional layers \citep{kipf2016semi}, GraphSAGE \citep{hamilton2017inductive}, edge convolutions \citep{gilmer2017neural}, and graph attention \citep{velivckovic2018graph}. In their basic formulation, GNNs layers require graph connectivity to be provided as input to the model. Mini-batches of data can be formed by sampling several graphs from a pool of graphs or sampling sub-graphs with a fixed number of nodes or connections from the original graph \citep{zou2019layer,cong2020minimal}. The common disadvantages of these methods are that they require a given or fixed input graph and the requirement of a sparse graph to make the approach computationally feasible, especially for large graphs. Recently, methods which do not assume given or fixed graph connectivity were proposed. Such methods construct the graph dynamically during training. \citet{wang2019dynamic} proposed Dynamic Graph CNNs (DGCNN) using KNN to construct the graph on-the-fly in the feature space of the neural network. Later, a graph learning model was proposed in \citet{cosmo2020latent} that builds a probability graph as a weighted adjacency matrix for an optimal classification result. Thus, the graph is built in a fully connected manner, implying that the model cannot exploit the possible sparseness of the graph. A more general Differential Graph Module (DGM) \citep{kazi2022differentiable} explicitly models sparse latent graphs employing the Gumbel-top-k trick \citep{kool2019stochastic}, overcoming the dense graph limitations by fixing the number of neighbours for each node which allows working with bigger graphs. We take inspiration from DGM for our sampling scheme, and we detail the major differences in Section \ref{sec:method}.
%The idea was further extended in PGC-DGCNN \citep{tran2018filter}, employing the shortest path connections between the neighbours to increase distant neighbours' contributions.
\subsection{Node feature imputation in GNNs}
\label{subsec:gnns_node_imputation}

GNNs models typically cannot deal with attribute-incomplete graph data directly, where rows represent nodes and columns feature channels. However, in real-world scenarios, features are often only observed for a subset of the nodes. Several works address missing node features in graph machine learning tasks. SAT \citep{chen2020learning} uses transformer-like models for feature completion followed by independent GNNs to solve the downstream task, which leads to a sub-optimal solution. GCNMF \citep{taguchi2021graph} overcomes this limitation by employing a Gaussian mixture model (GMM) to represent missing node features and jointly learns the GMM and GNNs parameters, however, this significantly increases the number of trainable parameters, implying high computational cost. PaGNNs introduced partial aggregation functions to propagate only the observed features \citep{jiang2020incomplete}. However, these cannot scale to large graphs \citep{rossi2021unreasonable}. Recently a discrete diffusion-based feature reconstructions framework was proposed, which leads to a simple, fast and scalable iterative algorithm \citep{rossi2021unreasonable}, though the method is designed to work for homophilic graphs.
\section{PROBLEM FORMULATION}
\label{sec:problem_formulation}

Let the matrix $\mathbf{X}=\{\mathbf{x}_{i}\}_{i=1}^{\mathrm{N}}$ denote a $\mathrm{d}$-dimensional dataset and $\mathbf{Y}$ represents its corresponding target vector. Without loss of generality, we assume each data vector $\mathbf{x}_{i}$ contains numerical variables referred to  $\mathrm{d} \in \{1,\dots,\mathrm{d}_{\mathrm{n}}\} $ and categorical variables indexed by $ \mathrm{d} \in \{\mathrm{d}_{\mathrm{n+1}},\dots,\mathrm{d}_{\mathrm{n}}+\mathrm{d}_{\mathrm{c}}\} $ with $\mathrm{d}_{\mathrm{n}}+\mathrm{d}_{\mathrm{c}}=\mathrm{d}$. We assume that each $\mathrm{d}_{\mathrm{th}}$ categorical variable takes values among $\mathrm{C}_{\mathrm{d}_{\mathrm{th}}}$ classes. The dataset $\mathbf{X}$ is referred as mixed if $\mathrm{d}_{\mathrm{n}}>0 ,\mathrm{d}_{\mathrm{c}}>0$, numerical if $\mathrm{d}_{\mathrm{c}}=0$ and categorical when $\mathrm{d}_{\mathrm{n}}=0$. By definition some percentage $\mathrm{I}_{\mathrm{init}} \in [0,1)$ of dataset entries $\mathrm{x}_{ij} \in \mathbf{X}$ are missing (corrupted). We associate a binary matrix $\mathbf{M} \in \{0,1\}^{\mathrm{N} \times \mathrm{N}}$ to identify missing and observed variables, where $1$ corresponds to the observed values, and $0$ indicates missing ones. Note that the corruption process can be of many types: MCAR, MNAR, MAR, and it is generally unknown to the user. The imputation process aims to provide a plausible estimation $\hat{\mathrm{x}}_{ij}$ for unobserved values $\mathrm{x}_{ij}$, such that (i) the imputed dataset $\widehat{\mathbf{X}}$ would be as close as possible to the real complete dataset (if such exists), (ii) the imputed dataset $\widehat{\mathbf{X}}$ has to achieve strong downstream task performance if adopting $\widehat{\mathbf{X}}$ as input to predict the corresponding target vector $\mathbf{Y}$.

\paragraph{Dataset preprocessing} We assume that the dataset $\mathbf{X}$, by definition, is properly normalized and corrupted in advance, hence the corresponding corruption matrix $\mathbf{M}$ is determined. Training different imputation approaches discussed in this paper requires distinct data preprocessing strategies. A straightforward way is to employ statistics of observed values. In our work, numerical values are initially assessed with \emph{mean} statistics, while categorical ones are approximated with the corresponding \emph{most frequent class} unless otherwise stated. Some of the imputation baselines discussed in this paper also require one-hot encoding of categorical values. The preprocessed dataset is denoted as $\overline{\mathbf{X}}$, which is the input to the subsequent EGG-GAE module.

\vspace{-0.5em}
\section{METHOD} 
\label{sec:method}

We propose a general solution for the tabular data MDI problem based on graph representation learning, whose overall pipeline is depicted in Fig. \ref{fig:model_pipline}. At every iteration, a mini-batch of data is sampled and preprocessed, according to the procedure described in Section \ref{subsec:minibatch_preprocessing}. This mini-batching is necessary, as it allows the algorithm to scale to large datasets. Next, we build a graph where each node is a row of the sampled mini-batch, and its corresponding node features are a non-linear mapping of the original inputs. The connectivity between nodes is learned through a differentiable sampling procedure, instead of being fixed as in previous works (e.g., \citep{spinelli2020missing}). We describe different variations of this last component in Section \ref{subsec:architecture}. We also propose two extensions to this basic architecture, i.e., ensembling and what we call \textit{prototype nodes}, in Section \ref{subsec:extensions}. The entire system is trained end-to-end with a combination of imputation losses and a downstream classification loss, as described in Section \ref{subsec:objective}.

\subsection{Mini-batch preprocessing} 
\label{subsec:minibatch_preprocessing}

Like in \citet{spinelli2020missing}, we turn MDI into a predictive task by employing a surrogate task, we randomly remove elements from the mini-batch and impute them to enable our model to reconstruct missing values. In contrast to denoising auto-encoders \citep{vincent2008extracting}, we only predict the removed elements rather than reconstructing the entire input. In order to obtain a surrogate batch-level corruption matrix $\mathrm{M}$ we use the MCAR mechanism to mask a certain percentage $\mathrm{I}_{\mathrm{b}} \in [0, 1 - \mathrm{I}_{\mathrm{init}})$ of the sampled batch $\mathrm{X} \sim \overline{\mathbf{X}}$. Precomputed statistics replace numerical masked values, while unobserved categorical entries are replaced with auxiliary tokens. We replace each categorical variable with a trainable dense embedding, whose size is a hyperparameter. Note that initial missing values, which have to be imputed during the inference, are represented as observed variables in the surrogate batch-level corruption matrix $\mathrm{M}$. The concatenation of the preprocessed batch parts: categorical $\mathrm{X}^{\mathrm{c}}$ and numerical $\mathrm{X}^{\mathrm{n}}$, yields the final preprocessed batch. In order to avoid cumbersome notation, we refer to the final batch representation as $\mathrm{X}$.

\subsection{Architecture}
\label{subsec:architecture}

\begin{figure}[tp]
    %\vspace{.3in}
    \centerline{\includegraphics[width=5cm]{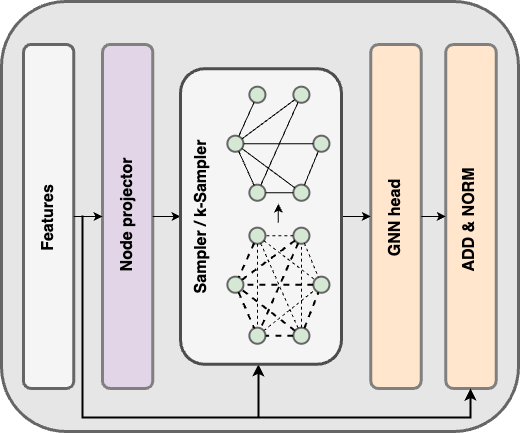}}
    %\vspace{.3in}
    \caption{Schematic depiction of the EGG module.}
    \label{fig:EGGNN}
\end{figure}

The core feature of the proposed EGG-GAE model is the EGG block, which endows the tabular representation of the sampled mini-batch with a graph topology in order to predict the missing values with a GNN module. The general EGG module comprises three components, as depicted schematically in Fig. \ref{fig:EGGNN}: (i) a \textit{node projector} transforms input features $\mathrm{H}_{\mathrm{k}}$ by projecting each row into a new space $\mathrm{H}^{\mathrm{g}}_{\mathrm{k}}$ that we call the graph embedding space; (ii) A \textit{sampler}, which obtains the edge set $\mathcal{E}_{\mathrm{k}}$ by sparsifying all possible edges between nodes; (iii) a GNN head that operates on the obtained graph $\mathcal{G}=(\mathrm{H}_{\mathrm{k}},\mathcal{E}_{\mathrm{k}})$. The EGG blocks can be stacked subsequently while their outputs concatenated to obtain the final representation $\mathrm{H}_{\mathrm{out}}$. In the following, we describe two variations of EGG blocks: the standard EGG blocks sample each edge independently, and a restricted $k$-EGG block that samples exactly $k$ neighbors for each node.

Before the first EGG block, the preprocessed batch $\mathrm{X}$ is encoded via an initial mapping function:
\begin{equation}
    \mathrm{H}_{0}=\mathrm{MLP}_{\mathrm{FP}}(\mathrm{X})
    \label{FeaturePropagationEq}
\end{equation}
\noindent where $\mathrm{MLP}_{\mathrm{FP}}$ is a two-layer MLP with a ReLU activation and batch normalization in between.

\paragraph{EGG}
Each EGG block first projects the input  $\mathrm{H}_{\mathrm{k}}$ with an additional row-wise operation, obtaining:
\begin{equation}
    \mathrm{H}^{\mathrm{g}}_{\mathrm{k}}=\mathrm{MLP}_{\mathrm{k}}(\mathrm{H}_{\mathrm{k}})
    \label{NodeProjector}
\end{equation}
\noindent where $\mathrm{MLP}_{\mathrm{k}}$ has the same architecture as $\mathrm{MLP}_{\mathrm{FP}}$. The sampler block represents pairwise edge relations of the projected $\mathrm{H}^{\mathrm{g}}_{\mathrm{k}}$ by first forming a probability matrix $\mathrm{P}$ where the $i$-th, $j$-th element is computed as:
\begin{equation} 
\mathrm{P}_{ij}=\exp^{||{\mathrm{h}}_{i} - {\mathrm{h}}_{j}||^{2}}
\label{prob_matrix}
\end{equation}
\noindent where ${\mathrm{h}}_{i}, {\mathrm{h}}_{j}$ are $i$-th and $j$-th rows of matrix $\mathrm{H}^{\mathrm{g}}_{\mathrm{k}}$. Each element of $\mathrm{P}_{ij}$ represents the probability of sampling edge $(i,j)$ in the output graph. In order to sample an undirected graph, corresponding to a strictly upper triangular adjacency matrix $\Tilde{\mathrm{A}}_{\mathrm{k}}$, we combine a Gumbel-Softmax trick \citep{jang2016categorical} for sampling from $\mathrm{P}$ with masking: 
\begin{equation}
    \Tilde{\mathrm{A}}_{ij} = (1+\mathrm{exp}(\nicefrac{(\log{(\mathrm{P}_{ij} \odot \mathrm{J}_{ij})} + \mathrm{G}_{ij} \odot \mathrm{J}_{ij})}{\tau}))^{-1}
    \label{GumbelSigmoid}
\end{equation}
\noindent where $\tau$ is a separate temperature hyperparameter, $\mathrm{J}$ is a strictly upper triangular matrix with ones above the main diagonal, $\odot$ is the Hadamard product, and $\mathrm{G}_{ij} \sim \text{Gumbel}(0,1)$. We obtain a sparse matrix by then thresholding $\Tilde{\mathrm{A}}_{\mathrm{k}}$ at $0.5$. The final sparse unweighted adjacency matrix $\mathrm{A}_{\mathrm{k}}=$ is computed as:
\begin{equation} 
\mathrm{A}_{\mathrm{k}}=\Tilde{\mathrm{A}}_{\mathrm{k}}+\Tilde{\mathrm{A}}_{\mathrm{k}}^{\mathrm{T}}+\mathrm{I}
\label{adj_symm}
\end{equation}
\noindent where $\mathrm{I}$ is the identity matrix. In order to have a valid gradient for back-propagation with respect to the thresholding operation, we use a straight-through estimator in the backward pass \citep{jang2016categorical} to allow the gradient flow through the sampling scheme. The final input of the GCN head is then a sparse unweighted adjacency matrix $\mathrm{A}_{\mathrm{k}}$ along with the corresponding feature representation $\mathrm{H}_{\mathrm{k}}$. The updated node representation $\mathrm{H}_{\mathrm{k+1}}$ is computed as: 
\begin{align}
    \hat{\mathrm{H}}_{\mathrm{k+1}}&= \mathrm{GCN}(\mathrm{H}_{\mathrm{k}}, \mathrm{A}_{\mathrm{k}})
    \label{gcn}
    \\
    \mathrm{H}_{\mathrm{k+1}}&= \mathrm{LayerNorm}(\hat{\mathrm{H}}_{\mathrm{k+1}} + \mathrm{H}_{\mathrm{k}})
    \label{gcn_norm}
\end{align}
where $\mathrm{GCN}$ is a graph convolutional layer \citep{kipf2016semi} or any other message-passing layer that operates on the graph connectivity.
\paragraph{$\boldsymbol{k}$-EGG} 
We also explore a variant of EGG, called $k$-EGG, which is more directly inspired to the sampling procedure in \citet{kool2019stochastic}. It forces a sparsity of the adjacency matrix by limiting the number of neighbours sampled per node to a fixed constant $k$. For each row $i$ of $\mathrm{A}_{\mathrm{k}}$, instead of thresholding each entry, we extract the first $k$ edges corresponding to the $k$ highest values, obtaining $\{(i,j_{1}),\dots,(i,j_{k})\}_{i=1}^{\mathrm{n}}$ and filling with ones the corresponding positions of matrix $\Tilde{\mathrm{A}}_{\mathrm{k}} \in \{0,1\}^{\mathrm{n} \times \mathrm{n}}$. The unweighted adjacency matrix $\mathrm{A}_{\mathrm{k}}$ is computed as in Eq. \eqref{adj_symm}.

\paragraph{Comparisons with DGM} Our graph sampling procedure is inspired to the recently proposed DGM block \citep{kazi2022differentiable}, with a number of important differences that we mention here. First, DGM was designed for a graph scenario where the full set of nodes is available beforehand and a \textit{single} underlying graph connectivity is assumed to exist. Instead, we work on randomly sampled mini-batches of data coming from a generic tabular dataset. Second, since the size of mini-batch is a user-defined parameter, we use directly the straight-through estimator in the backward pass, avoiding an additional surrogate loss as in \citet{kazi2022differentiable}. The proposed module does not require to limit the output dimensions of the node projector to small values, hence obtaining greater model expressivity. Finally, the GNN head in this paper follows a design with skip-connections and layer normalization.

\subsection{Extensions to the basic formulation}
\label{subsec:extensions}

We describe here two simple extensions of the model that we have found to work consistently better in our experimental evaluation.

\paragraph{Ensembling}
The proposed model has two sources of stochasticity: mini-batch sampling and continuous relaxation of discrete random variables (the edge sampling procedure). We propose to exploit this stochasticity during inference by sampling the batches from the test set until each datum (node) has the desired number of predictions, so that each node has multiple predictions relying on different neighbourhoods. In the classification case, we select the maximum average soft prediction, whereas we use the mean prediction for the regression case. 

\paragraph{Prototype nodes}
In this paper, we also propose to use trainable prototype nodes which we refer to as $\mathrm{H}_{\mathrm{pr}}$. Applying mini-batching to tabular or graph-structured data may lead in exceptional cases to a lack of data expressiveness during the forward pass due to the same sources of stochasticity mentioned above. For instance, in the case of tabular data, there is no guarantee that  the sampled mini-batch composes a reliable set of neighbours for the particular datum prediction. Therefore, prototype nodes are designed to encode common data patterns and allow each data point to have reliable neighbours regardless of the sampled mini-batch. The number of prototypes nodes is a hyperparameter. We initialize the prototypes nodes randomly. However, other strategies such as data mean can be applied. The prototype nodes are then added to $\mathrm{H}_{k}$ before every EGG block. $\mathrm{H}_{\mathrm{pr}}$ does not participate explicitly in the objective function while contributing as sampled neighbours.

\subsection{Objective}
\label{subsec:objective}

The objective function consists of two main terms: downstream loss and imputation loss. The downstream loss $\mathcal{L}_{\mathrm{task}}$ paired with the model construction allows to perform end-to-end training, including sampling scheme optimization (in contrast to the DGM paper \citep{kazi2022differentiable}). In the experiments, we use classification task datasets. Therefore, the proposed model is optimized through a cross-entropy loss. 

The imputed data $\widehat{\mathrm{X}}$ along with the network predictions $\widehat{\mathrm{Y}}$ are obtained through representation $\mathrm{H}_{\mathrm{out}}$ as:
\begin{align}
    \widehat{\mathrm{X}}&=[\mathbf{H}^{\mathrm{n}}(\mathrm{H}_{\mathrm{out}}),
    \mathbf{H}^{\mathrm{c}}_{i}(\mathrm{H}_{\mathrm{out}})]
    \label{imp_heads}\\
    \widehat{\mathrm{Y}}&=\mathbf{H}^{\mathrm{task}}(\mathrm{H}_{\mathrm{out}})
    \label{task_head}
\end{align}
\noindent where $\mathbf{H}^{\mathrm{n}}, \{\mathbf{H}^{\mathrm{c}}_{i}\}_{i=1}^{\mathrm{c}}, \mathbf{H}^{\mathrm{task}}$ are linear projectors, acting row-wise on the input matrix $\mathrm{H}_{\mathrm{out}}$.
The downstream task loss $\mathcal{L}_{\mathrm{task}}$ depends on the dataset, in our case that is cross entropy. Note that downstream task loss implicitly contributes to finding the best solution for the imputation problem and latent graph representation learning. The prototype nodes do not participate in any losses explicitly. However, implicit contribution through neighbour batch nodes allows the gradient to flow backwards and learn their representation. The mini-batch masking procedure introduces a surrogate objective to simulate the presence of missing data for which the reconstruction loss is computed. We optimize the MDI solution with the sum of Eq. \eqref{imp_num_loss} and \eqref{imp_cat_loss}, $\mathrm{M}^{\mathrm{n}},\mathrm{M}^{\mathrm{c}}$ are numerical and categorical parts of surrogate batch-level corruption matrix $\mathrm{M}$ corresponding to $\mathrm{X}^{\mathrm{n}},\mathrm{X}^{\mathrm{c}}$ subsequently.

\begin{align} 
    \mathcal{L}^{\mathrm{n}}_{\mathrm{imp}}&=
    \frac{\mathrm{MSE}(\mathrm{X}^{\mathrm{n}} \odot \mathrm{M}^{\mathrm{n}}, \widehat{\mathrm{X}}^{\mathrm{n}}\odot \mathrm{M}^{\mathrm{n}})}{\sum \mathrm{M}^{\mathrm{n}}}  
    \label{imp_num_loss}
    \\
    \mathcal{L}^{\mathrm{c}}_{\mathrm{imp}}&=\frac{\mathrm{CrossEntropy}(\mathrm{X}^{\mathrm{c}} \odot \mathrm{M}^{\mathrm{c}}, \widehat{\mathrm{X}}^{\mathrm{c}} \odot \mathrm{M}^{\mathrm{c}})}{\sum \mathrm{M}^{\mathrm{c}}}
    \label{imp_cat_loss}  
    \\
    % \mathcal{L}_{\mathrm{imp}}&=\frac{d_{n}}{d}\mathcal{L}^{\mathrm{n}}_{\mathrm{imp}} + \frac{d_{c}}{d}\mathcal{L}^{\mathrm{c}}_{\mathrm{imp}}
    % \label{imp_loss}
    % \\
    % \mathcal{L}_{\mathrm{task}}&=\mathrm{CrossEntropy}(\mathrm{Y}, \widehat{\mathrm{Y}})
    % \label{task_loss}
    % \\
    \mathcal{L}_{\mathrm{h}}&=\sum_{i=1}^{\mathrm{N}}\sum_{j=1}^{\mathrm{N}}\bar{\mathrm{a}}_{ij}{\mathrm{a}}_{ij}
    \label{homophilic_loss}
\end{align}
Graph learning generally assumes a homophilic structure of the data, i.e., similar patterns are connected with a higher probability, which in our case means that we expect patterns of the same class to be linked. We can explicitly enforce homophily by penalising interclass entries within the sampled adjacency matrix $\mathrm{A}$. The target vector $\mathbf{Y}$ provides an idealized adjacency matrix $\mathrm{A}_{\mathbf{Y}}$, where $\mathrm{A}_{\mathbf{Y}_{ij}}=1$ if $\mathbf{Y}_{i}$ and $\mathbf{Y}_{j}$ belong to the same class, otherwise zero.  The complement of the idealized adjacency is denoted as $\bar{\mathrm{A}}_{\mathbf{Y}}$. Equation \eqref{homophilic_loss} represents the penalization term, where $\bar{\mathrm{a}}_{ij} \in \bar{\mathrm{A}}_{\mathbf{Y}}$ and $\mathrm{\mathrm{a}}_{ij} \in \mathrm{A}$. The final objective $\mathcal{L}$ is computed as a weighted combination of losses (\ref{imp_num_loss}-\ref{homophilic_loss}):
\begin{equation} 
\mathcal{L}=\alpha \mathcal{L}_{\mathrm{task}} + \beta (\mathcal{L}^{\mathrm{n}}_{\mathrm{imp}} + \mathcal{L}^{\mathrm{c}}_{\mathrm{imp}})+ \gamma \mathcal{L}_{\mathrm{h}}
\label{full_loss}
\end{equation}
\noindent where $\alpha$, $\beta$, $\gamma$ are loss weights.
\section{EXPERIMENTS AND RESULTS}
The primary experimental comparison is discussed in Section \ref{subsec:imputation_exp}, whereas in Section \ref{ablation_exp}, we concentrate on architectural ablations. All experiments are conducted at the following noise levels: $10\%, 20\%, 30\%, 40\%$, and $50\%$.

\paragraph{Datasets}
We validate our method on 15 datasets from the UCI repository by artificially introducing missing values using MCAR, MNAR, or MAR mechanisms (the setup for MAR and MNAR is replicated from \cite{muzellec2020missing}). Table \ref{table:dataset_stat} displays the aggregate statistics for the datasets. The training and test parts of the SUSY dataset were subsampled so that they could be compared to the majority of imputation baselines. The remaining datasets are divided into training sets (70\%) and validation sets (30\%). To satisfy a real-world scenario, we optimise the model with respect to additional noise introduced into the validation set (using the MCAR mechanism regardless of the source of initial missingness) and report the results concerning the initially missing values. Following the relevant literature, we evaluate the imputation and post-imputation prediction performance for each experiment.

\begin{table}[t]
\footnotesize
\caption{Dataset statistics. For each dataset, we provide the number of samples, the number of numerical features, and the number of categorical datasets.} \label{table:dataset_stat}
\begin{center}

\scalebox{0.8}{
\begin{tabular}{|c|c|c|c|c|}
\hline
{Data type} & {Dataset} & {\#Samples} & {\#Num.Fs.} & {\#Cat.Fs.}\\
\hline
\multirow[c]{7}{*}{\rotatebox[origin=c]{90} {\textbf{Numerical}}} & Yeast & 1484 & 8 & 0 \\ 
     & Wireless & 2000 & 7 & 0 \\
     & Abalone & 4177 & 8 & 0 \\
     & Wine-quality & 4898 & 11 & 0  \\
     & Page blocks & 5473 & 10 & 0 \\
     & Electrical grid stability & 10000 & 14 & 0\\
     & SUSY (small) & 25000 & 18 & 0 \\

\hline
\multirow[c]{3}{*}{\rotatebox[origin=c]{90} {\textbf{Mixed}}} & Anuran & 7195 & 22 & 3 \\ 
    & Default credit card & 30000 & 14 & 10 \\ 
    & Adult & 32561 & 6 & 8 \\
\hline
\multirow[c]{5}{*}{\rotatebox[origin=c]{90} {\textbf{Categorical}}} & Car & 1728 & 0 & 6 \\
    & Phishing websites & 2456 & 0 & 9 \\
    & Letter & 20000 & 0 & 16 \\
    & Chess & 28056 & 0 & 6 \\
    & Connect & 67557 & 0 & 42 \\
\hline
\end{tabular}
}
\end{center}
\end{table}

% \begin{table}[h]
% \caption{Sample Table Title} \label{sample-table}
% \begin{center}
% \begin{tabular}{ll}
% \textbf{PART}  &\textbf{DESCRIPTION} \\
% \hline \\
% Dendrite         &Input terminal \\
% Axon             &Output terminal \\
% Soma             &Cell body (contains cell nucleus) \\
% \end{tabular}
% \end{center}
% \end{table}

\paragraph{Algorithms}
In the experiments the proposed architectures EGG-GAE and $k$-EGG-GAE, are compared with two statistical imputation methods: Mean \citep{little2019statistical}, KNN \citep{troyanskaya2001missing}, two machine learning imputation approaches: MICE \citep{van2011mice}, MissForest (MF) \citep{stekhoven2012missforest} and four deep learning approaches: MIDA \citep{gondara2018mida}, GINN \citep{spinelli2020missing}, GAIN \citep{yoon2018gain}, and NN (the proposed architecture wherein EGG block is substituted with an MLP one that is identical to $MLP_{\mathrm{FP}}$).
To provide a fair comparison with the baselines, we apply the hyper-parameters and data preprocessing steps from the original papers for all datasets. The proposed models and NN baseline utilize the data pipeline described in this paper.

\begin{figure*}[t]
  \centering
  \begin{subfigure}[b]{0.45\linewidth}
    \centerline{\includegraphics[width=\textwidth]{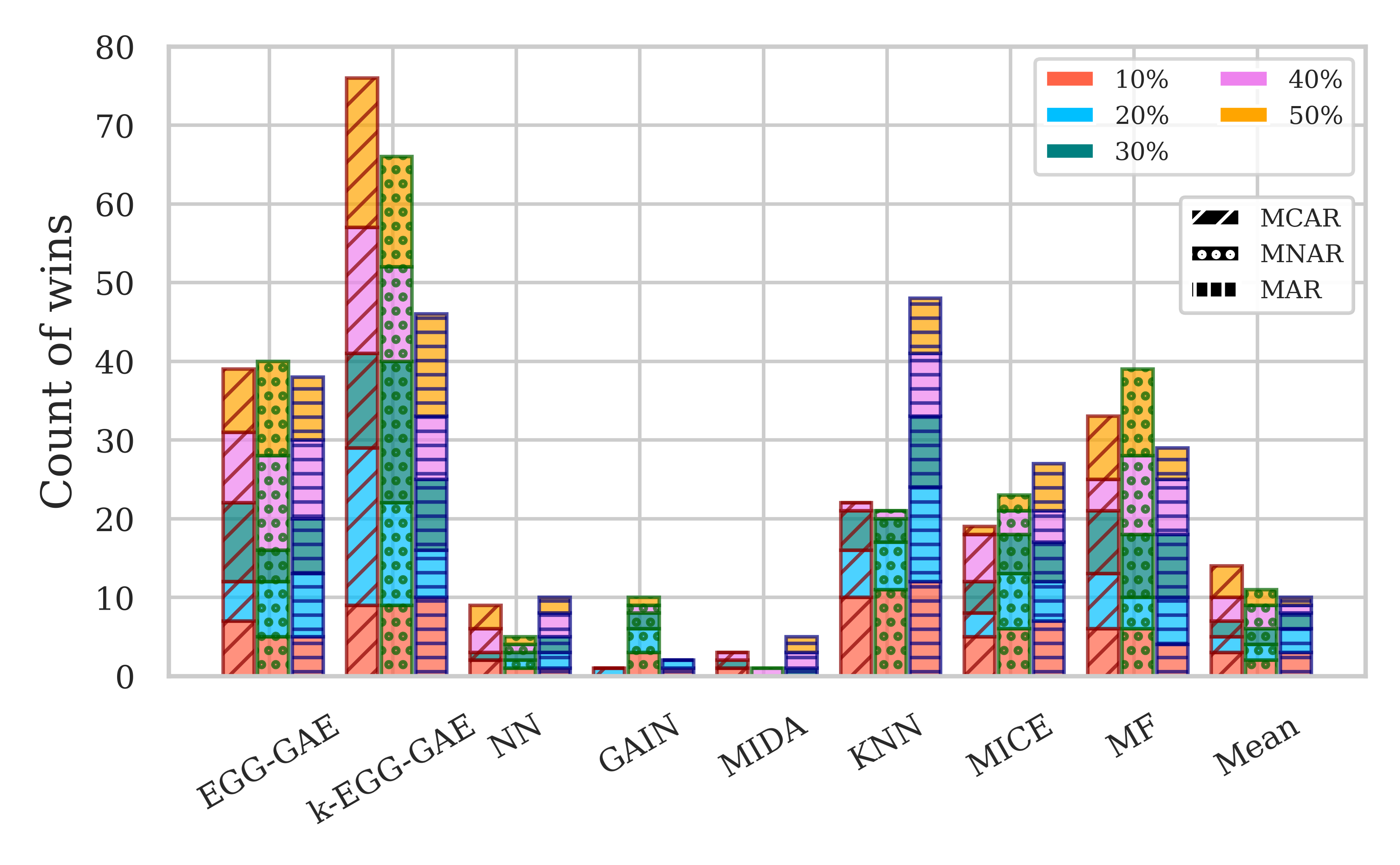}}
    \caption{Unified count of wins}
    \label{fig:imputatuion_counts}
  \end{subfigure}
  \begin{subfigure}[b]{0.45\linewidth}
    \centerline{\includegraphics[width=\textwidth]{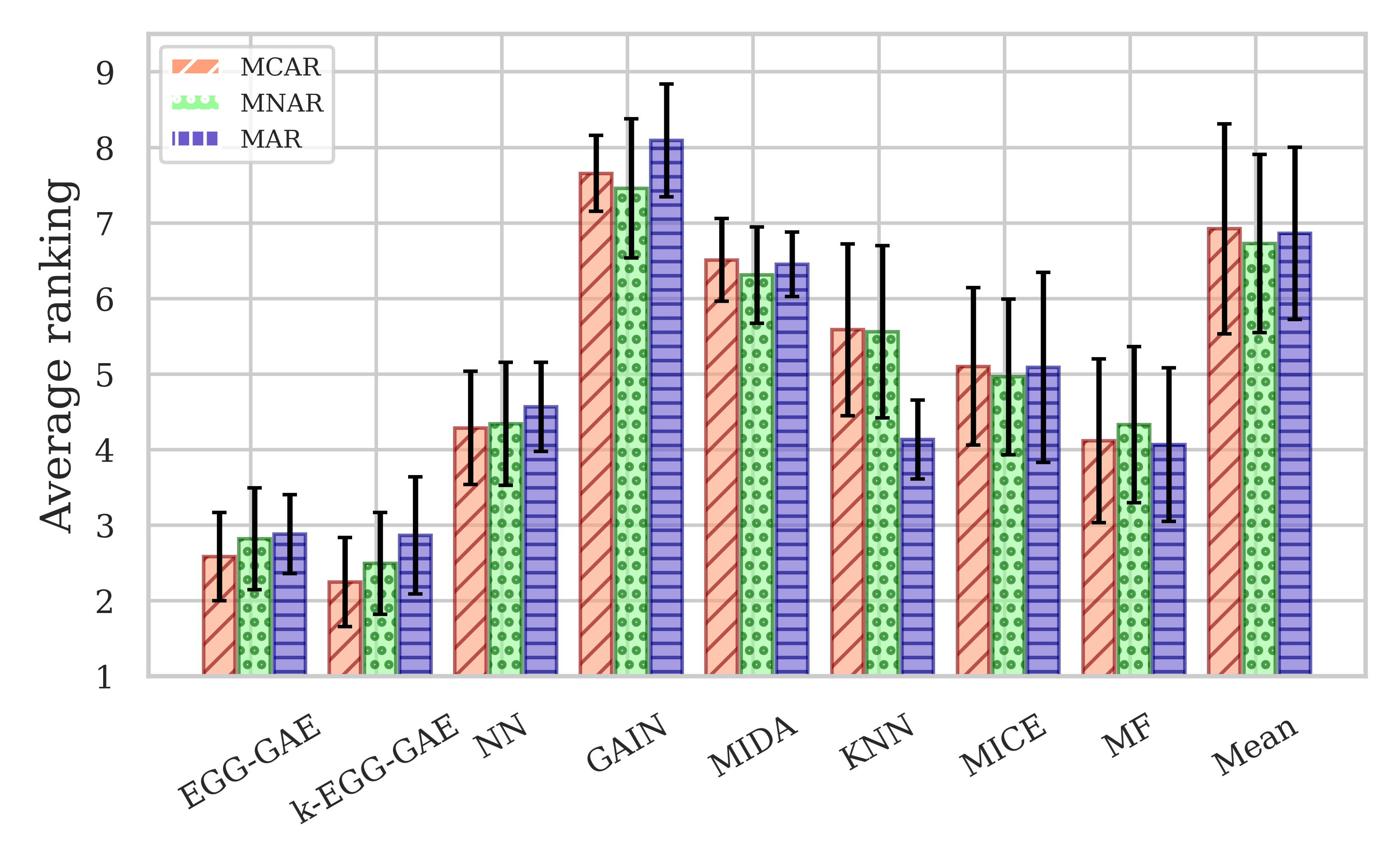}}
    \caption{Unified average ranking}
    \label{fig:imputation_AR}
  \end{subfigure}
  \caption{Unified average ranking computed for the MCAR, MNAR, and MAR scenarios.}
  \label{fig:unified_plots}
\end{figure*}

\paragraph{Proposed architecture details}
The surrogate batch level corruption is introduced with the MCAR mechanism and $\mathrm{I}_{b}=0.2$. The number of EGG blocks is equal to $1$. The hidden representations of all MLPs (feature propagation block and node mapper) are equal to $300$. The batch size is equal to $300$. The regularization homophily parameter is equal to $0.1$. The  temperature parameter $\tau$ linearly decrease from $0.5$ to $0.01$. During inference, the ensembling parameter is equal to $5$. EGG-GAE and $k$-EGG-GAE share the majority of architectural hyperparameters, and for $k$-EGG-GAE we fix the number of sampled neighbours per node at $k=5$. We utilize RMSprop for the optimization with the learning rate equal to $0.0001$.

\vspace{-0.5em}
\subsection{Imputation} 
\label{subsec:imputation_exp}

The main set of experiments addresses the imputation reconstruction and predictive performance of the proposed networks in comparison to baseline algorithms utilising the MCAR, MNAR, and MAR mechanisms. The predictive performance of an MDI solution for tabular data is typically measured by classical machine learning (ML) algorithms for a downstream task. To assess post-imputation downstream task performance we employ random forest \cite{breiman2001random}. We show a schematic result with a unified count of wins and a unified average ranking that takes all levels of noise into account.

The unified count of wins shows the summary for each level of noise and missing mechanism (MCAR, MNAR and MAR). It represents the unified number of times that each method achieves the best performance with respect to the imputation or post-imputation task metrics, i.e., the lowest for RMSE, MAE and the highest in imputation and predictive accuracy. To compute the average ranking we first rank the model for each dataset according to the performance metric. The average ranking is then calculated by averaging the obtained rankings across all datasets. We obtain a separate average ranking for each level of noise, resulting in a matrix consisting of average rankings for every level of noise. The unified average ranking represents multiple average rankings based on a variety of performance metrics and is displayed as a bar graph with error bars. The bar height represents the mean of average rankings regarding the performance metrics and noise levels, while the error bars show corresponding variations. Note that GINN does not participate in the unified count of wins or unified average ranking, because execution time of GINN for big datasets is unfeasible within a reasonable amount of time. The full results for the scenario in which $20\%$ of the entries are missing can be found in the supplementary material, Appendix \ref{sec:imputation_sup},  where GINN model is represented as well.

In Fig. \ref{fig:unified_plots}, it is evident that the suggested models outperform the baselines for every missing mechanism, especially for higher noise levels. Fig.  \ref{fig:imputatuion_counts} shows that $k$-EGG-GAE achieves the best score considerably more often than EGG-GAE, especially for MCAR and MNAR scenarios. Aggregating the results across every noise level, we observe that the EGG-GAE and $k$-EGG-GAE together accumulate $53\%, 49\%, 39\%$ of the best cases against the $15\%, 18\%$, and $22\%$ of its best competitor for the MCAR, MNAR and MAR mechanisms, respectively. The machine learning baselines (KNN, MICE, and MF) are the strongest competitors, achieving the best performance in $33\%, 40\%$, and $48\%$  of the cases (cumulatively) for the MCAR, MNAR, and MAR scenarios, accordingly. In Fig.\ref{fig:imputation_AR}, we can see that the proposed model framework using MLP instead of EGG (referred to as NN) performs just as well as MF on average, even though it rarely achieves the highest score. In addition, we can see that the proposed EGG-GAE and $k$-EGG-GAE models stay roughly on par.

\subsection{Ablation experiments}
\label{ablation_exp}

We perform ablation studies over the numerical datasets (reported in Table \ref{table:dataset_stat}). The ablation study examines model architecture alterations, evaluating ensembling and prototype nodes. The results are averaged across five runs and presented as unified average rankings based on end-to-end accuracy, RMSE and MAE. In Section \ref{time_comparison_sec.} we analyse the proposed methods further by comparing training/inference timings for various neural baselines.The architectural experiments can be found in the suplementary materials, Appendix \ref{sup_arch_exp}. 
\begin{figure*}[t]
  \centering
  \begin{subfigure}[b]{0.45\linewidth}
    \centerline{\includegraphics[width=\textwidth]{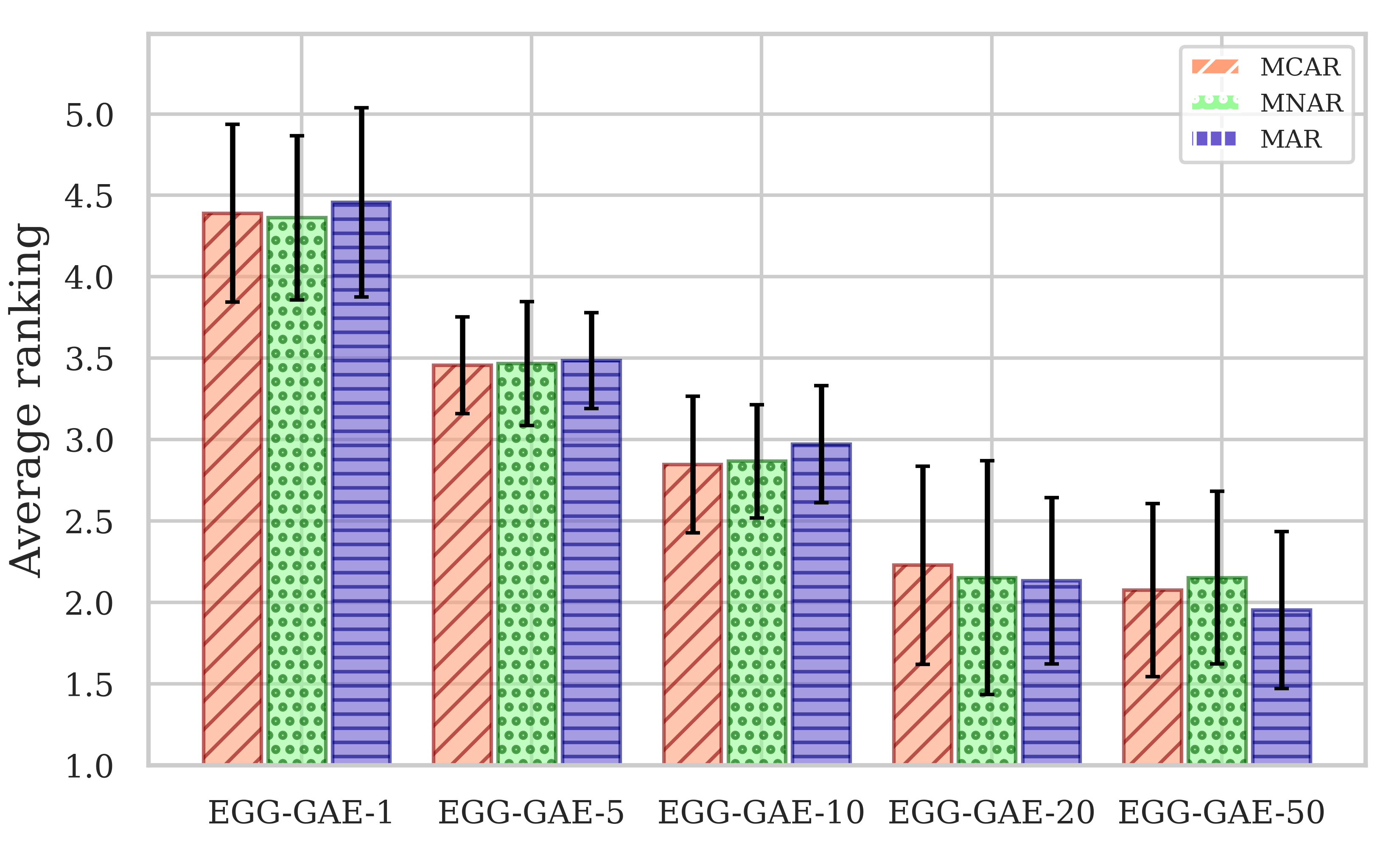}}
    \caption{Ensembling}
    \label{ensemble_average_ranking}
  \end{subfigure}
  \begin{subfigure}[b]{0.45\linewidth}
    \centerline{\includegraphics[width=\textwidth]{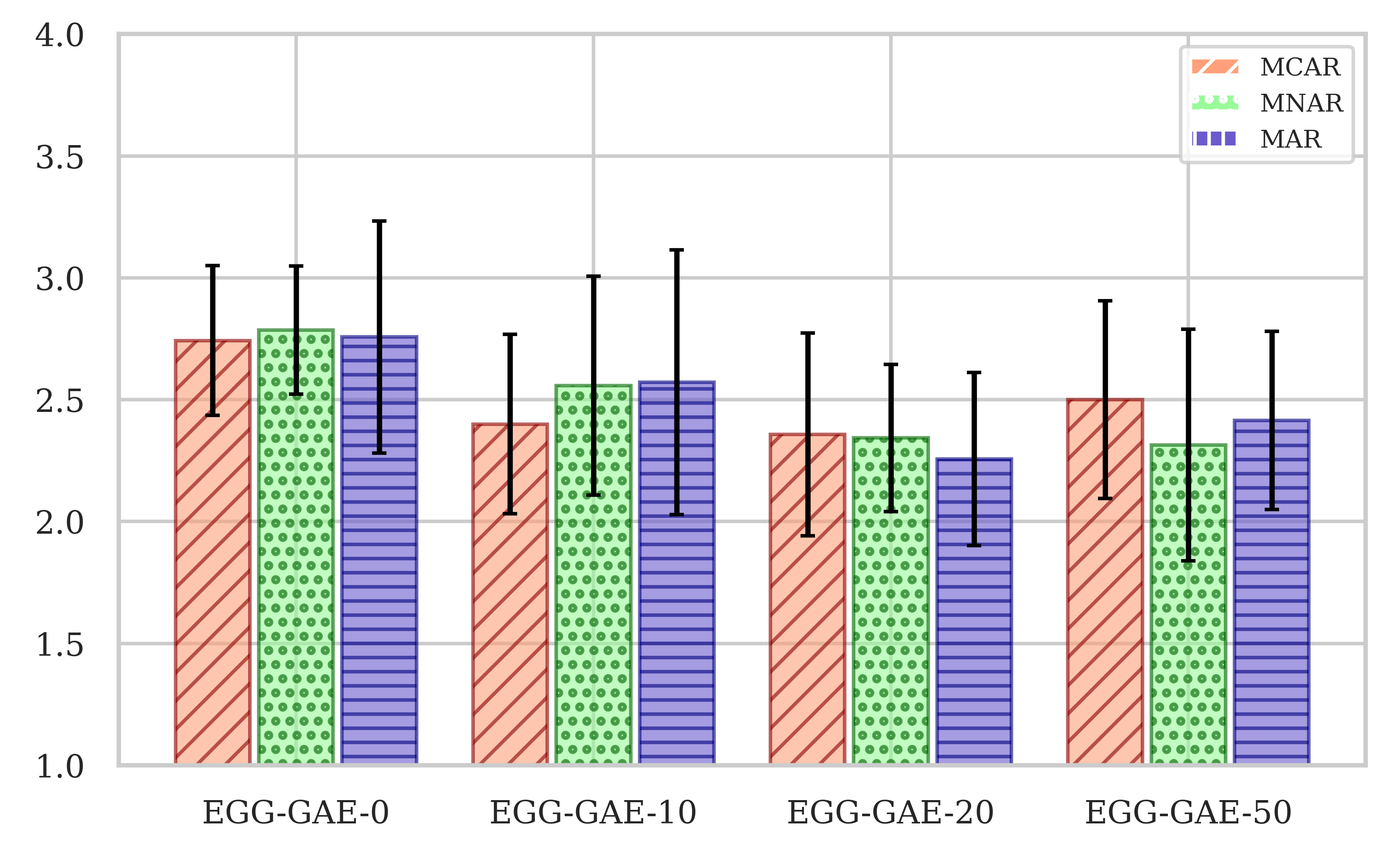}}
    \caption{Prototype nodes}
    \label{proto_average_ranking}
  \end{subfigure}
  \caption{Unified average ranking computed for the MCAR, MNAR, and MAR scenarios. A value following the model's name indicates the variable parameter: a number of prototype nodes or a number ensembling iterations.}
  \label{fig:ens_proto}
\end{figure*}

Figure \ref{ensemble_average_ranking} shows that performing ensembling during inference enhances the performance of downstream and MDI problems regardless of missingness mechanisms. We can see that increasing the number of ensembling iterations on average causes improvement for every type of missingness, as hypothesized. Both EGG-GAE-20 and EGG-GAE-50 have close mean values, while the corresponding variations share the same interval (around $[1.3, 2.8]$); this suggests that the plateau is reached when the number of iterations surpasses 20. We argue that performing ensembling during the inference leads to (i) reliable performance (ii) enhancing the performance of downstream and MDI problems. 
Figure \ref{proto_average_ranking} demonstrates that the models with learnable prototype nodes (EGG-GAE-10, EGG-GAE-20, and EGG-GAE-50) have a lower mean average ranking compared to the model without them (EGG-GAE-0), independent of the missingness mechanisms scenario. Increasing the number of prototype nodes helps to achieve better performance. However, we can see that introducing a significant number of prototype nodes might worsen the performance (EGG-GAE-50). We believe that  the number of prototype nodes should be at least equal to the number of classes of downstream tasks. However, adding too many prototype nodes can impair the performance. We believe it affects the sampling scheme by increasing the likelihood that the graphs will be built primarily with prototype nodes, resulting in a poorer solution.

\vspace{-1em}
\subsubsection{Time comparison} 
\label{time_comparison_sec.}

\begin{figure}[t]
  \begin{subfigure}[b]{0.45\columnwidth}
    \includegraphics[width=\linewidth]{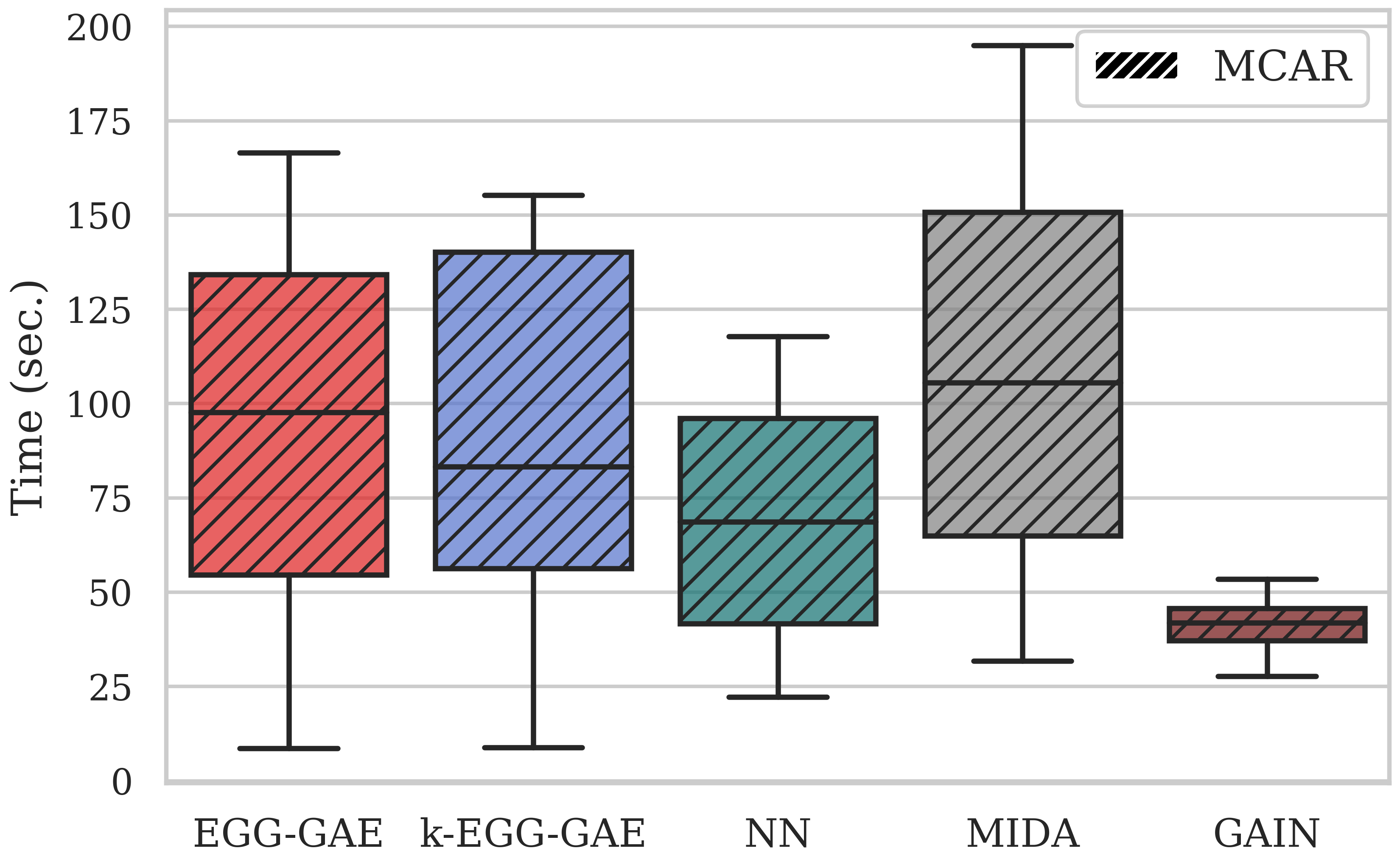}
    \caption{Average training time}
    \label{fig:time_ATT}
  \end{subfigure}
  \hfill %%
  \begin{subfigure}[b]{0.45\columnwidth}
    \includegraphics[width=\linewidth]{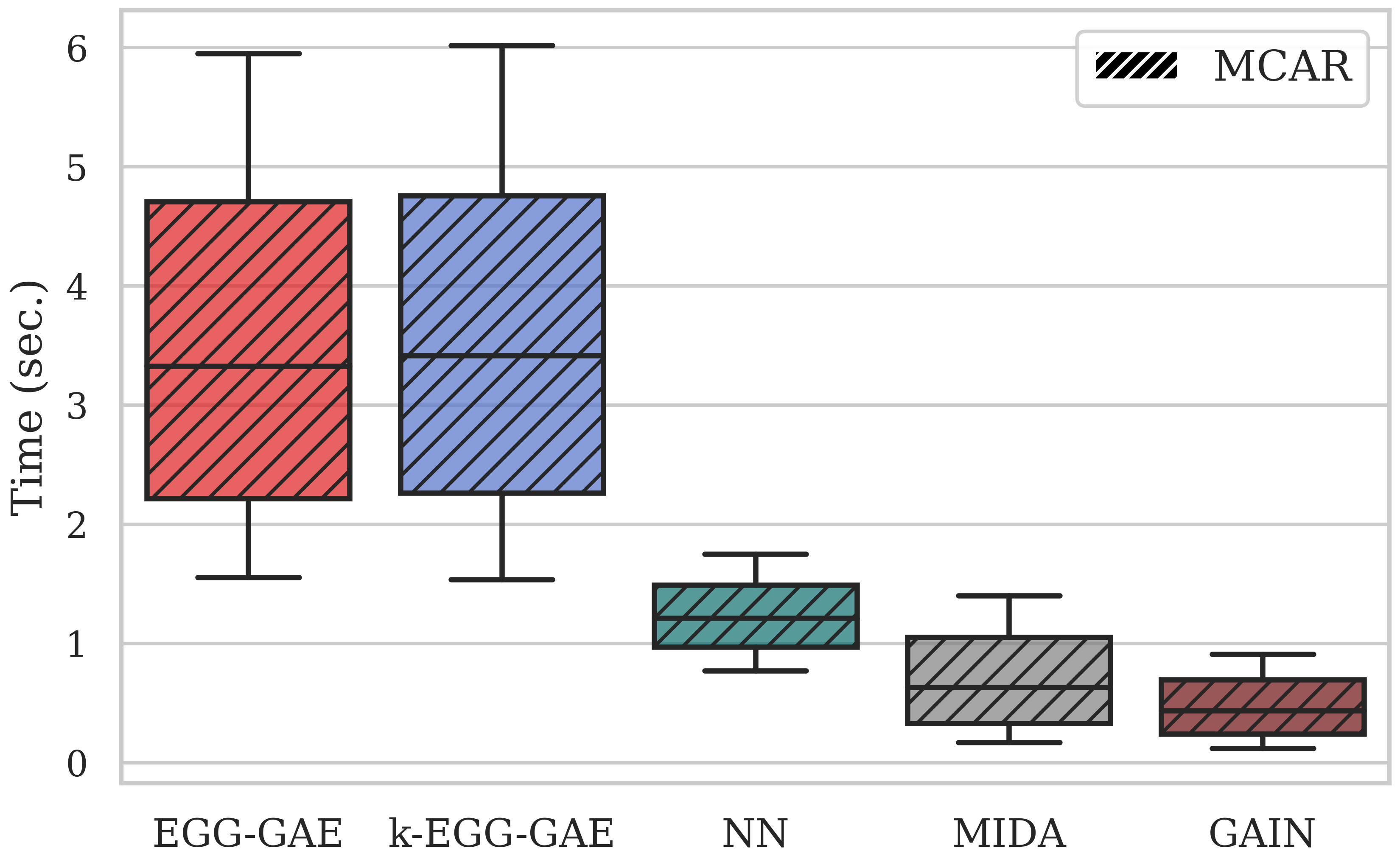}
    \caption{Average inference time}
    \label{fig:time_AIT}
  \end{subfigure}
  
  \vskip 0.3cm 
  
  \begin{subfigure}[b]{0.45\columnwidth}
    \includegraphics[width=\linewidth]{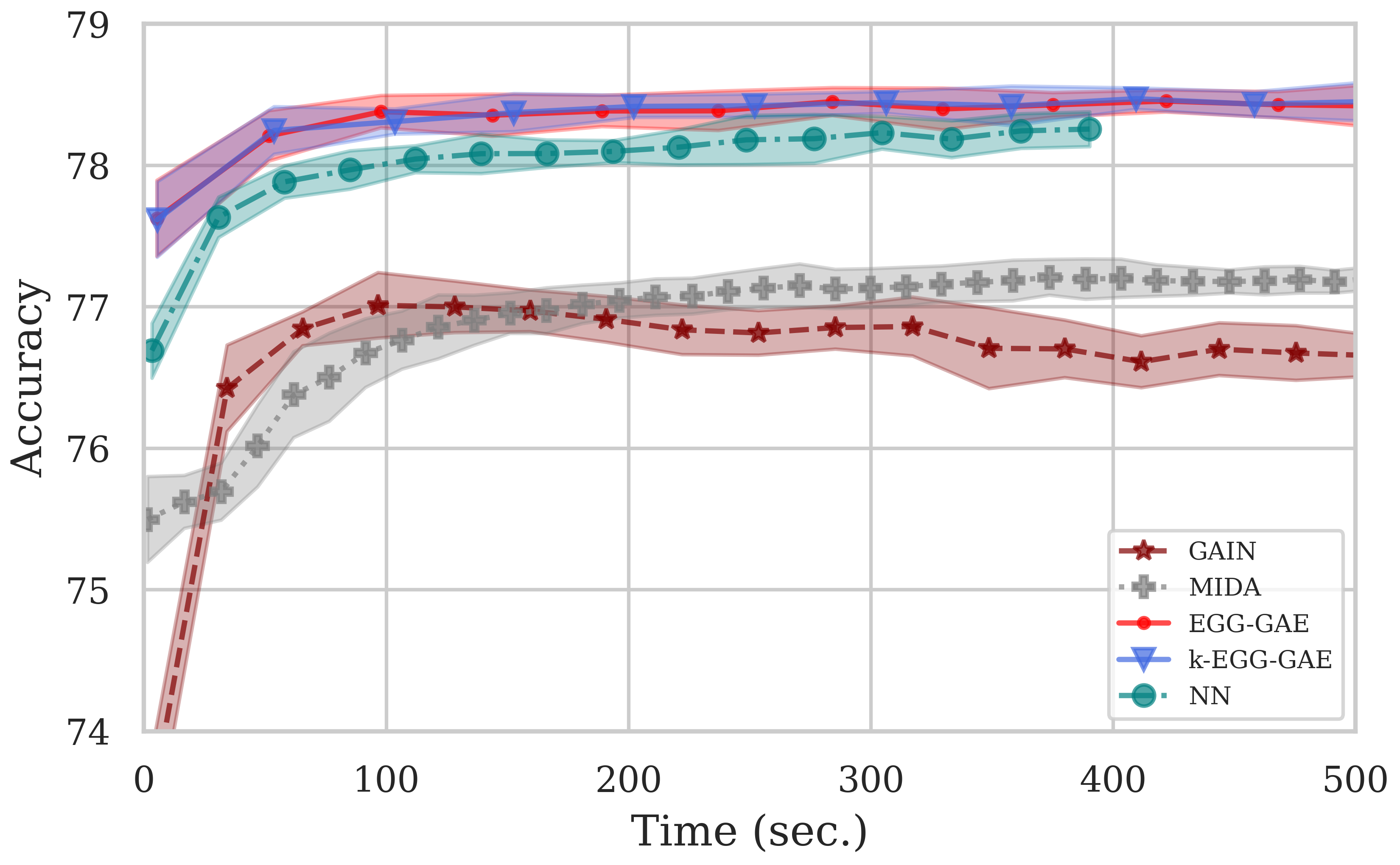}
    \caption{Accuracy over time}
    \label{fig:acc_over_time}
  \end{subfigure}
  \hfill %%
  \begin{subfigure}[b]{0.45\columnwidth}
    \includegraphics[width=\linewidth]{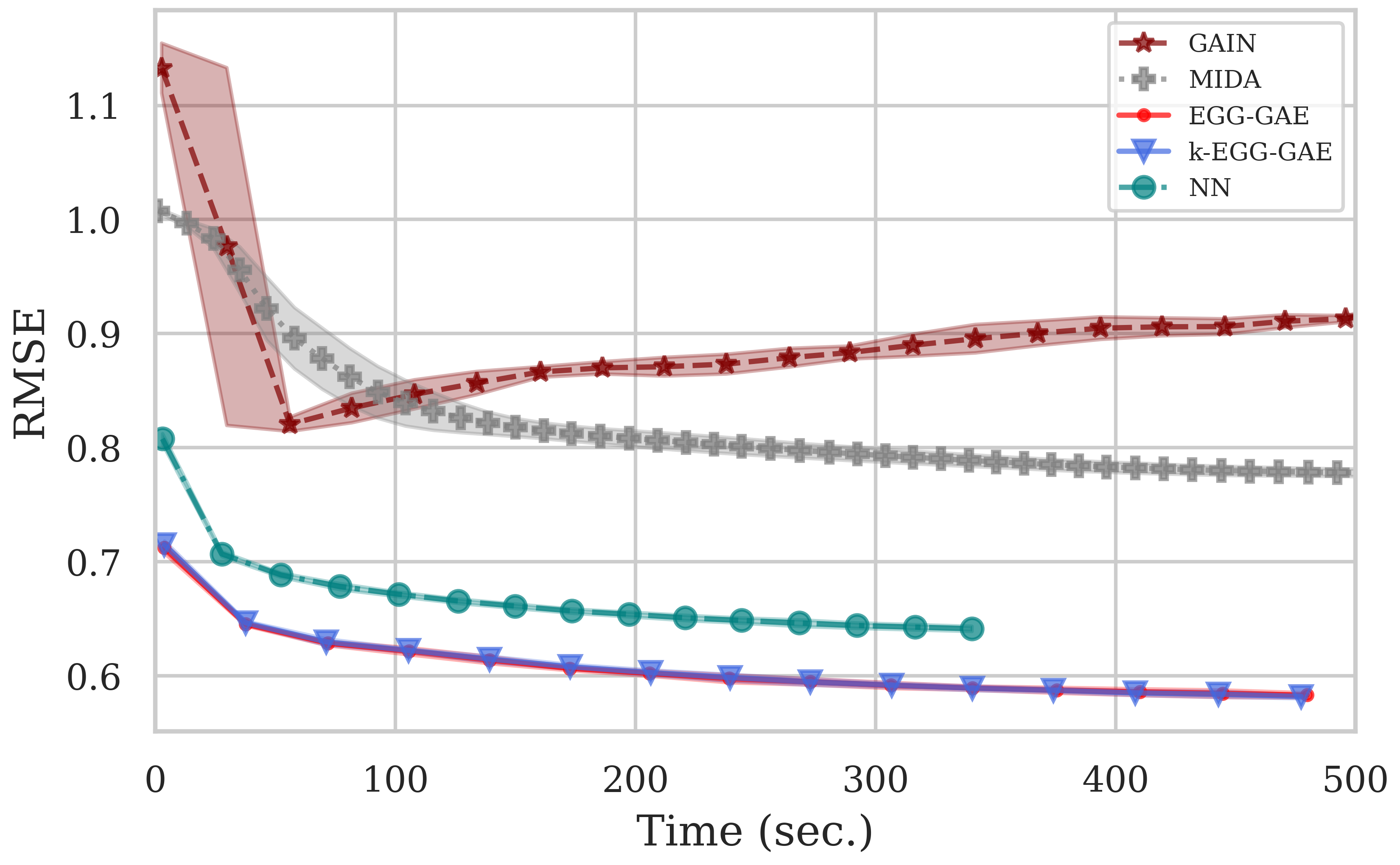}
    \caption{RMSE over time}
    \label{fig:rmse_over_time}
  \end{subfigure}
  \caption{Average training/inference and performace time comparison for the neural approaches.}
  \label{fig:time}
  
\end{figure}

Figure \ref{fig:time_ATT} represents the average training time until convergence of the validation loss for the numerical datasets, while Figure \ref{fig:time_AIT} depicts the average models inference time. Note that the proposed architecture EGG-GAE does not vary a lot between average time training along with average inference time. This follows from the batch size (300) that was used to train EGG-GAE and $k$-EGG-GAE, which is fixed for all experiments. Increasing the batch size will result in quadratic time consumption growth due to the pairwise distance calculation. The average training time is approximately the same as MIDA and  five times slower compared to GAIN. Average inference time is approximately 5-6 times greater compared to NN, MIDA and GAIN models, mostly due to the ensembling procedure. In Figures \ref{fig:acc_over_time} and \ref{fig:rmse_over_time} we illustrate the evolution of accuracy and RMSE throughout training time in seconds for SUSY, averaged over 10 runs.

% Experiments move to supplementary
%\input{experiments/heads_exp.tex}
%\input{experiments/homo_exp}
%\input{experiments/triplet_exp}

%over five levels of noise ($10\%, 20\%, 30\%, 40\%, 50\%$)
\section{DISCUSSION AND CONCLUSION}

In this paper, we propose a generic framework for handling missing values in tabular data, employing graph deep learning. Particularly, we presented an end-to-end trainable graph autoencoder (EGG-GAE) model for learning the underlying graph representation of tabular data applied to the MDI problem. We performed extensive experiments with real-world datasets (from different fields) and determined that our model outperformed current state-of-the-art algorithms in terms of imputation and downstream task performance. We described several improvements to our model by demonstrating that ensembling improves MDI and dataset task performances; we further introduced novel learnable {prototype nodes} to encode common data patterns and serve as a generic, reliable subset of nodes for the predicted graphs. Finally, we introduced a regularization method that forces homophily in the learned latent graph representation.

As future works, the proposed EGG-GAE network can be applied to any type of data to introduce a graph topology for, e.g., imputing missing data over images, audio, or other types of high-dimensional data, exploiting the modularity of modern deep learning architectures. In addition, the Euclidean distance calculation can be substituted with a trainable network to construct probabilistic graphs based on a learned metric distance function. The assumption and limitations of the proposed framework are described in supplementary materials, Appendix \ref{sec:assumptions_and_limitations}.

% \newpage
% \bibliographystyle{apalike}
% \bibliography{biblio.bib}

\onecolumn
\aistatstitle{Supplementary Materials for \textit{EGG-GAE: scalable graph neural networks for tabular data imputation}}
\appendix
\section{ARCHITECTURAL EXPERIMENTS}
We perform architectural experiments over the numerical datasets (reported in Table \ref{table:dataset_stat}). The results are averaged across five runs and presented as unified average rankings based on end-to-end accuracy, RMSE and MAE. In Section \ref{homo_experiment_sec.} we analyse the proposed homophily penalization term, evaluate different GNN heads in Sec. \ref{heads_experiment_sec.}. The effect of extra manipulation of the embedding space obtained by the node projector is investigated in Sec. \ref{triplet_experiment sec.}. Examine the impact of varying the number of neighbours sampled per node in the restricted sampling scheme of the $k$-EGG-GAE model in Section \ref{restricted_sampling_sec.}.
\label{sup_arch_exp}
\subsection{Homophily experiment} 
\label{homo_experiment_sec.}
\begin{figure}[h] 
    \centering
    \includegraphics[width=0.5\textwidth]{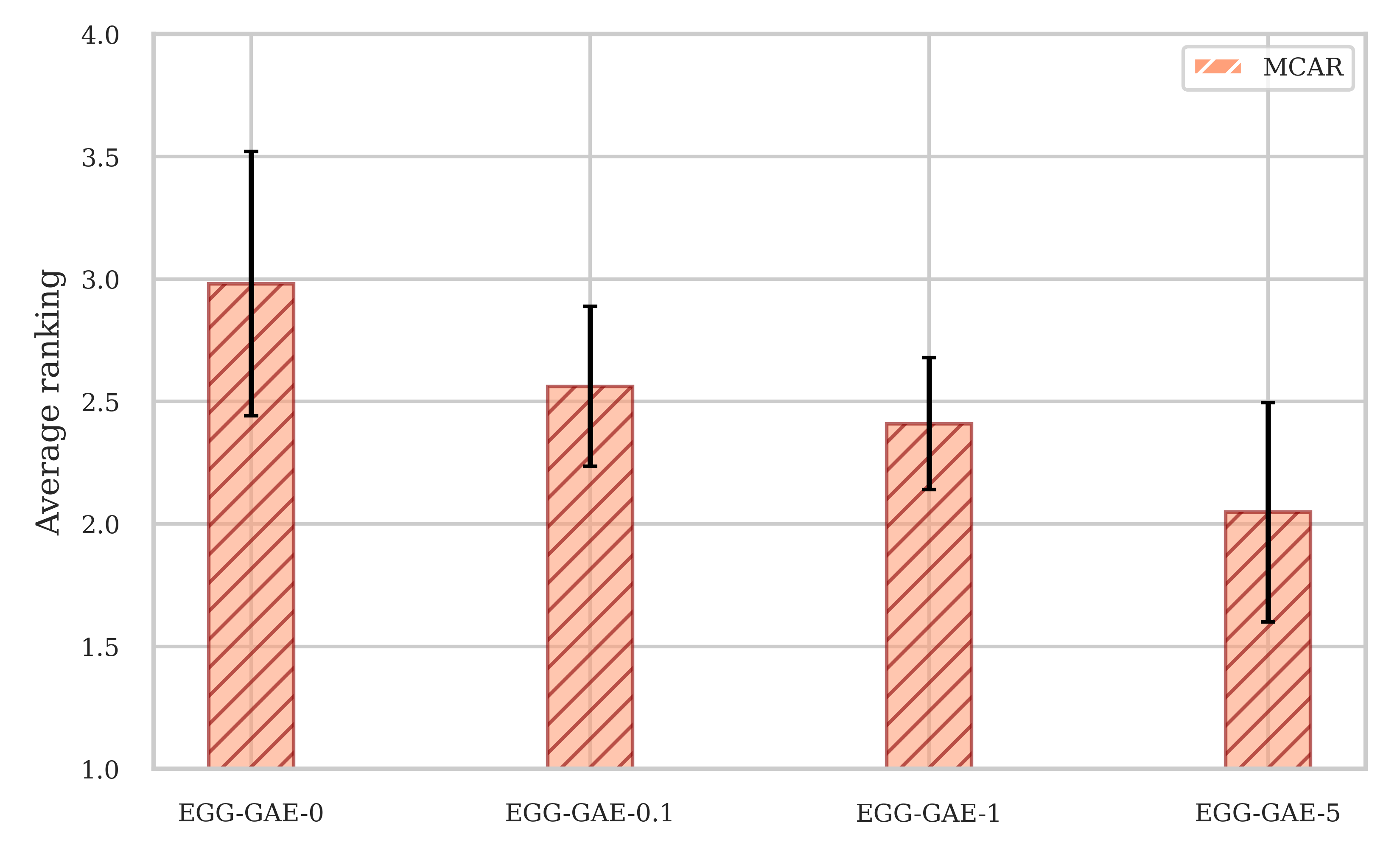}
    \caption{Unified average ranking computed for the MCAR scenario. A value following the model name indicates the regularization hyperparameter $\gamma$.}
    \label{homo_average_ranking}
\end{figure}

We argue that boosting $\gamma$ in Eq.\ref{full_loss} enhances the sampling scheme of the EGG-GAE model by restricting the sampling of non-homophilic neighbours. We further investigate the influence of the proposed homophily loss adapted to EGG-GAE model. Fig. \ref{homo_average_ranking} demonstrates that, on average, using the homophily regularisation term is beneficial. Increasing the regularisation hyperparameter $\gamma$ results in an improved unified solution on average. Although high penalization improves the performance, the variation of EGG-GAE-$5$ indicates that the performance enhancement has plateaued.

\subsection{Heads experiment} 
\label{heads_experiment_sec.}

\begin{figure}[h] 
    \centerline{\includegraphics[width=0.5\textwidth]{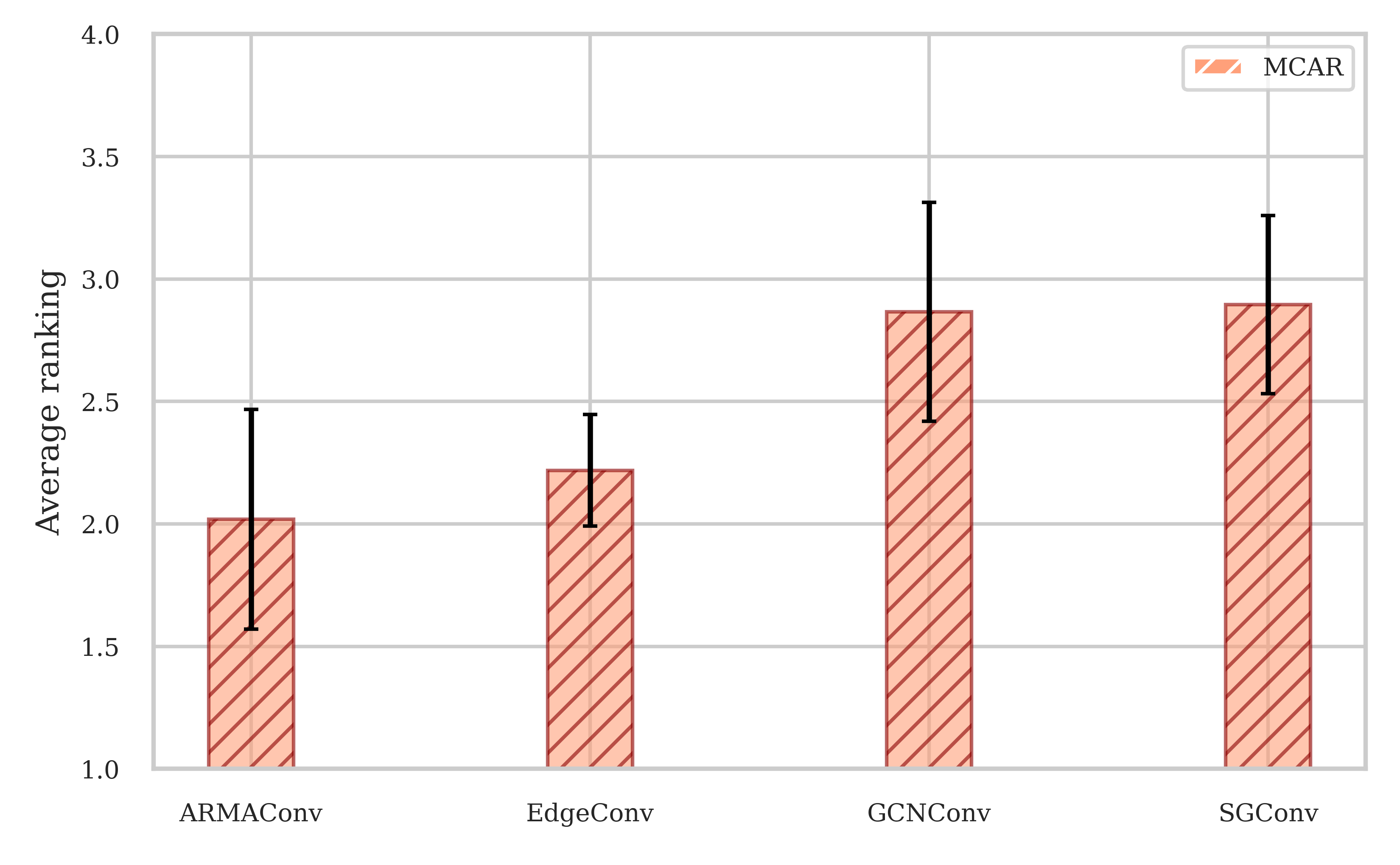}}
    \caption{Unified average ranking computed for the MCAR scenario. The model name indicates the head of EGG-GAE model.}
    \label{heads_average_ranking}
\end{figure}

Here we inspect the performance change under different GNNs heads. We explore four heads: GCN \citep{kipf2016semi}, EdgeConv \citep{wang2019dynamic}, ARMAConv \citep{bianchi2021graph} and SGConv \citep{wu2019simplifying}. As can be seen in Fig. \ref{heads_average_ranking}, ARMAConv and EdgeConv on average perform better than GCNConv and SGConv, which achieve roughly the same results, further improving the results from Section \ref{subsec:imputation_exp}. We hypothesize a potential explanation of ArmaConv and EdgeConv superior performance compared to GCN and SGConv as follows. ArmaConv is more resistant to noise, which increases its resilience to incorrectly sampled connectivity, while EdgeConv intrinsically weights the contribution of each neighbour, providing additional noise resistance and reducing the contribution of not similar examples (which were sampled due to stochasticity) for concrete datum prediction. As a result, an additional filter is applied to the sampled nodes.

\subsection{Metric learning experiment} 
\label{triplet_experiment sec.}

\begin{figure}[h] 
    \centerline{\includegraphics[width=0.5\textwidth]{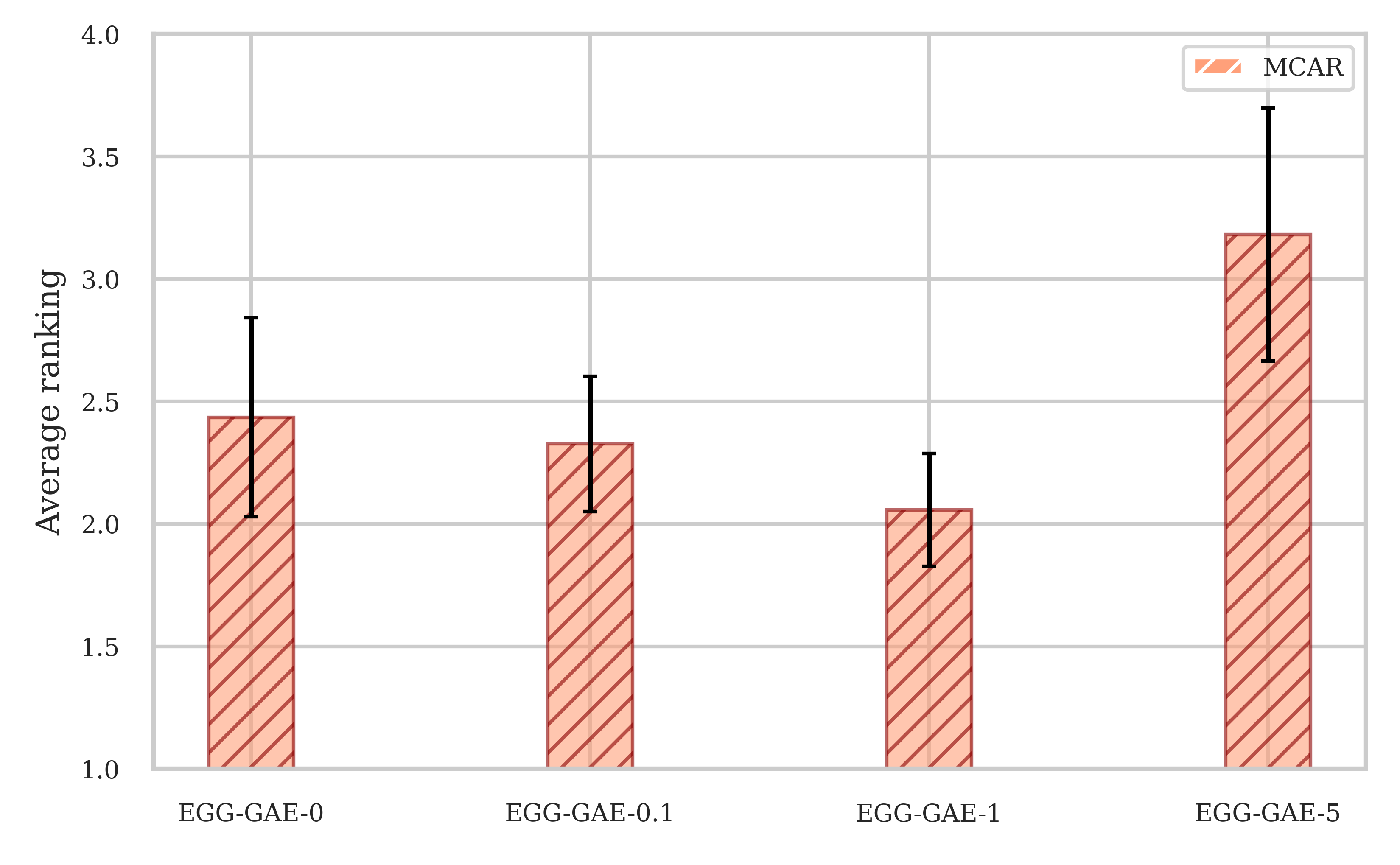}}
    \caption{Unified average ranking computed for the MCAR scenario. A value following the model name indicates the regularization hyperparameter $\eta$.}
    \label{triplet_average_ranking}
\end{figure}

In this part, we investigate the possibility of influencing the embedding space acquired by the node projector. We add additional regularization on the embeddings $\mathrm{H}^{g}_{\mathrm{k}}$ obtained by Eq. \ref{NodeProjector} using triplet loss \cite{schroff2015facenet} which is calculated as:
\begin{equation}
    \mathcal{L}_{t} =\eta \sum_{i}^{T}{\left[ \|{\mathrm{h}}^{a}_{i} - {\mathrm{h}}^{+}_{i}\|^{2} + \| {\mathrm{h}}^{a}_{i} - {\mathrm{h}}^{-}_{i}\|^{2} + \mathrm{m} \right]}
    \label{FeaturePropagationEq}
\end{equation}
\noindent where ${\mathrm{m}}$ is a margin and equal to $0.05$, $\eta$ is a regularization hyperparameter and $\{ {\mathrm{h}^{\mathrm{a}}_{i}}, {\mathrm{h}^{+}_{i}}, {\mathrm{h}^{-}_{i}}\}_{i}^{\mathrm{T}}$ are the triplets formed from embeddings ${\mathrm{H}^{\mathrm{g}}_{\mathrm{k}}}$ forcing the homophily by selecting  ${\mathrm{h}^{\mathrm{a}}_{i}}$ and ${\mathrm{h}^{+}_{i}}$ from the same class and ${\mathrm{h}^{-}_{i}}$ from the other. We mine the triplets with distance weighted margin-based approach \citep{wu2017sampling}. Fig. \ref{triplet_average_ranking} demonstrates applying further regularization on node embedding space can lead to a better solution; nevertheless, the scale parameter $\eta$ has to be carefully chosen, since high values of $\eta$ result in suboptimal solutions.

\subsection{Restricted sampling} 
\label{restricted_sampling_sec.}

\begin{figure}[h] 
    \centering
    \includegraphics[width=0.5\textwidth]{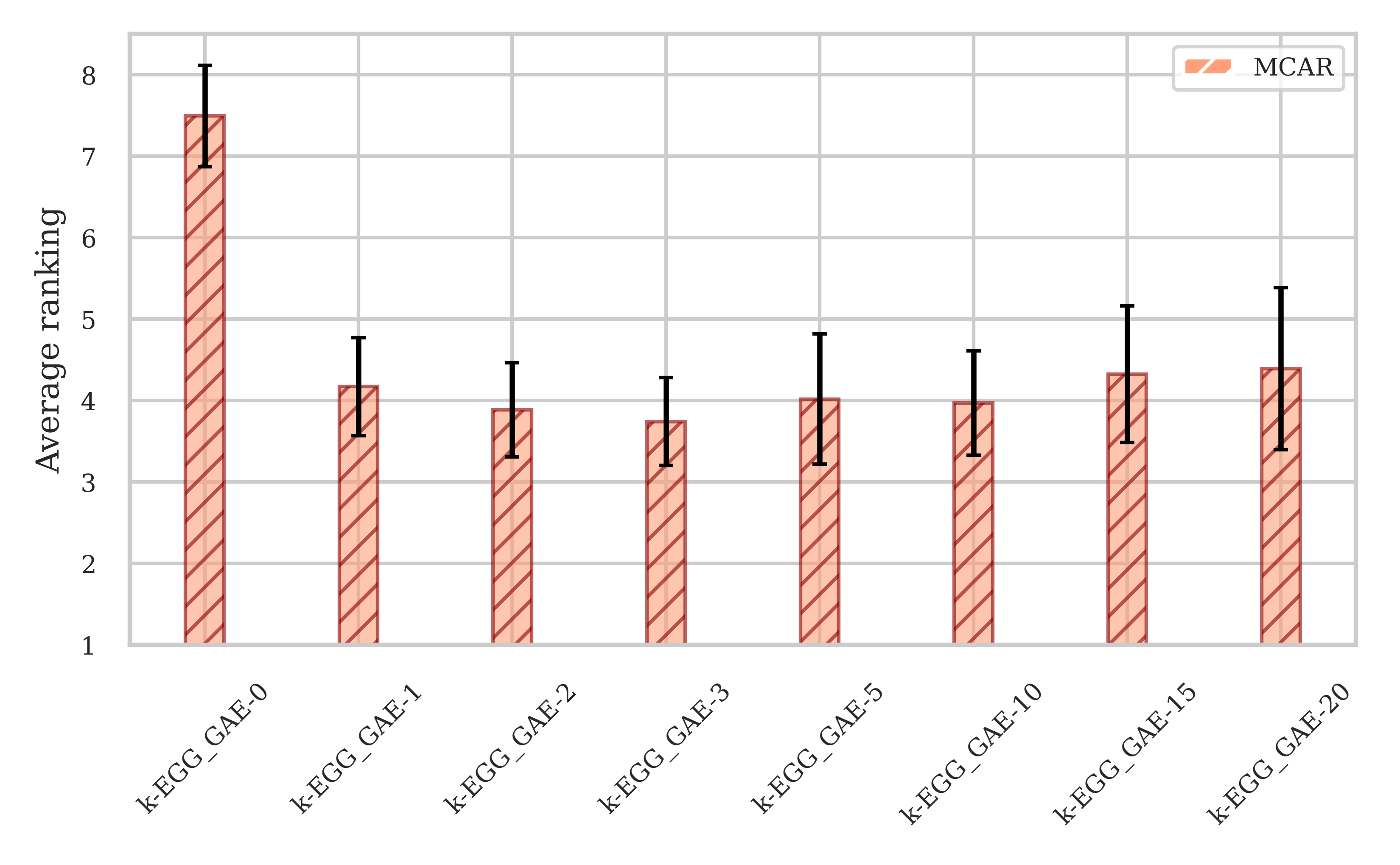}
    \caption{Unified average ranking computed for the MCAR scenario. A value following the model name indicates the number of neighbours sampled per node.}
    \label{kegg_vary_k_AR}
\end{figure}

In this section we investigate restrictive sapling procedure by varying the number of sampled neighbours k per node.  Figure \ref{kegg_vary_k_AR} demonstrates the corresponding experiment, where the model $k$-EGG-GAE-0 is a model which has only self-nodes. Models that rely on the sampled neighbourhood consistently outperform models with only self-nodes in terms of MDI solution and predictive accuracy. Next, we observe that increasing the number $k$ of sampled neighbours improves the performance on average, and that the optimal number of neighbours is $3$. In addition, as the number of sampled neighbours increases, both the average ranking and the variation increase. We hypothesise that this indicates that as the number of sampled neighbours $k$ increases, so does the proportion of noisy neighbours, which degrades performance.

\section{IMPUTATION EXPERIMENT}
\label{sec:imputation_sup}
The main set of experiments addresses the imputation reconstruction and predictive performance of the proposed networks in comparison to baseline algorithms utilising the MCAR, MNAR, and MAR mechanisms. The predictive performance of an MDI solution for tabular data is typically measured by classical machine learning (ML) algorithms for a downstream task. To assess post-imputation downstream task performance we employ random forest for all models \cite{breiman2001random} and provide the findings in Table \ref{table:RF_accuracy}. Tables \ref{table:imp_rmse}, \ref{table:imp_mae} and \ref{table:imp_catacc} display the MDI reconstruction error in terms of RMSE, MAE, and accuracy for numerical and categorical values, respectively, when 20\% of values are missing. Due to the fact that the execution time exceeds 24 hours, some GINN results are unavailable and denoted by ``-'' in the table. 

According to Tables \ref{table:RF_accuracy}-\ref{table:imp_catacc} we can see that for the majority of datasets the proposed EGG-GAE and $k$-EGG-GAE prevail as the best or second best solution in terms of post-imputation predictive performance and MDI solution, regardless of the missingness mechanism. Tables \ref{table:RF_accuracy} and \ref{table:imp_catacc} demonstrates that for categorical data, algorithms inferring similar data points (EGG-GAE, $k$-EGG-GAE, GINN, and KNNI) achieve the best predictive and MDI performance. Regarding MDI performance, the cumulative number of wins of models employing similar datapoints is 23 out of 24 cases.
Additionally, it is noticeable in Table \ref{table:imp_rmse}, that $k$-EGG-GAE model dominates in 50\% of cases in total, and 60\% cases considering MNAR missigness mechanism. From Table \ref{table:imp_mae}, we can see that the machine learning algorithm MF achieves the best result in 10 out of 24 cases, compared to the cumulative win of EGG-GAE and $k$-EGG-GAE models (the cases when the proposed models shares first and second place): in 9 out of 24 instances; however, from the schematic representation (Figure \ref{fig:imputatuion_counts}), it is evident that $k$-EGG-GAE dominates over the MF algorithm when all performance metrics are considered (predictive and imputation accuracy, RMSE, MAE).
\begin{table}[h]
\begin{center}

\caption{Post-imputation downstream task predictive performance in terms of accuracy (Average±Std, averaged over 5 runs) under MCAR, MNAR and MAR assumptions (the case in which $20\%$ of entries are missing). The best results for each dataset are highlighted in bold font, and the second-best result is underlined.}
\begin{adjustbox}{width=\textwidth}
%\resizebox{!}{.5\textwidth}{
\begin{tabular}{|c|c|c| c c c c c c c c c c| }%# *{13}{c}
\hline
{\textbf{Miss.}} & \textbf{Dataset type}& {\textbf{Dataset}} & {\textbf{EGG-GAE}} & {\textbf{k-EGG-GAE}} & {\textbf{NN}} & {\textbf{GINN}} & {\textbf{GAIN}} & {\textbf{MIDA}} & {\textbf{KNNI}} & {\textbf{MICE}} & {\textbf{MF}} & {\textbf{Mean}} \\
\hline
\multirow[c]{15}{*}{\rotatebox[origin=c]{90}{\textbf{MCAR}}} & \multirow[c]{7}{*}{\rotatebox[origin=c]{90}{Numerical}} & Yeast & \underline{51.27±0.68} & \textbf{52.02±1.55} & 49.78±1.19 & 48.28±0.93 & 50.07±1.04 & 50.37±1.37 & 46.19±0.0 & 51.12±0.0 & 50.97±0.68 & 48.88±0.0 \\
& & Wireless & 95.33±0.58 & \underline{95.33±0.0} & 95.0±0.0 & 91.44±2.34 & 90.78±1.68 & 92.11±2.14 & 91.33±0.0 & \textbf{95.67±0.0} & 95.0±0.0 & 89.0±0.0 \\
& & Abalone & 59.86±0.4 & 59.81±0.28 & 59.7±0.49 & 56.51±0.24 & 58.53±0.32 & 59.17±0.42 & 59.17±0.0 & \textbf{60.93±0.0} & \underline{59.91±0.74} & 57.1±0.0 \\
& & Wine quality & 50.48±0.14 & 50.43±0.44 & 50.43±0.42 & 49.61±0.28 & \underline{50.84±0.57} & 50.11±0.77 & \textbf{50.88±0.0} & 50.2±0.0 & 50.34±0.24 & 49.52±0.0 \\
& & Page blocks & 93.91±0.0 & 93.75±0.19 & 93.71±0.07 & 93.02±0.14 & 93.95±0.07 & 93.26±0.07 & \textbf{94.28±0.0} & \underline{94.03±0.0} & 94.03±0.12 & 93.42±0.0 \\
& & Electrical grid stability & \underline{95.69±0.27} & \textbf{95.82±0.2} & 95.64±0.08 & 90.87±0.18 & 93.0±0.0 & 94.22±0.15 & 94.07±0.0 & 94.53±0.0 & 95.13±0.74 & 93.0±0.0 \\
& & SUSY (small) & \textbf{75.79±0.11} & \underline{75.73±0.15} & 75.55±0.15 & -- & 75.02±0.08 & 75.08±0.12 & 75.28±0.0 & 75.35±0.0 & 75.38±0.16 & 75.52±0.0 \\
\cline{2-13}
& \multirow[c]{5}{*}{\rotatebox[origin=c]{90}{Categorical}} & Car & 70.0±0.77 & \underline{70.38±0.38} & 69.23±0.67 & 70.0±0.0 & 69.87±0.22 & 69.74±0.22 & 70.0±0.0 & \textbf{70.38±0.0} & 70.0±0.0 & 68.85±0.0 \\
& & Phishing website & \underline{82.59±0.28} & \textbf{82.92±1.86} & 81.44±0.57 & 78.0±0.28 & 81.61±1.24 & 81.77±0.49 & 82.27±0.0 & 79.8±0.0 & 80.79±0.49 & 79.8±0.0 \\
& & Letter & \underline{48.03±0.71} & 47.56±1.1 & 45.59±0.95 & 40.63±0.24 & 42.93±0.29 & 43.73±0.07 & \textbf{52.63±0.0} & 46.0±0.0 & 46.04±0.12 & 43.23±0.0 \\
& & Chess & 25.32±0.22 & 25.27±0.65 & 25.24±0.07 & -- & \underline{26.48±0.07} & 26.17±0.27 & 26.35±0.0 & 25.23±0.0 & 25.47±0.26 & \textbf{27.06±0.0}\\
& & Connect & 65.92±0.05 & 65.92±0.03 & 65.96±0.01 & -- & 65.86±0.01 & 65.94±0.01 & 65.96±0.0 & \textbf{66.03±0.0} & \underline{66.03±0.03} & 65.86±0.0 \\
\cline{2-13}
& \multirow[c]{3}{*}{\rotatebox[origin=c]{90}{Mixed}} & Anuran & \underline{92.01±0.37} & \textbf{92.28±0.33} & 90.12±0.37 & 85.62±0.89 & 87.35±0.33 & 86.27±0.27 & 91.11±0.0 & 89.26±0.0 & 91.54±0.23 & 85.28±0.0 \\
& & Adult & \textbf{81.92±0.31} & \underline{81.83±0.37} & 81.4±0.2 & -- & 79.68±0.58 & 80.02±0.02 & 79.69±0.0 & 80.32±0.0 & 80.85±0.27 & 80.13±0.0 \\
& & Default credit card & \underline{80.54±0.09} & 80.53±0.06 & \textbf{80.59±0.09} & -- & 80.26±0.06 & 80.18±0.04 & 80.47±0.0 & 80.51±0.0 & 80.52±0.05 & 80.2±0.0 \\
\hline 
\multirow[c]{15}{*}{\rotatebox[origin=c]{90}{\textbf{MNAR}}} & \multirow[c]{7}{*}{\rotatebox[origin=c]{90}{Numerical}} & Yeast & \underline{49.15±0.87} & 46.37±1.57 & 46.37±0.87 & 48.61±0.25 & 48.07±1.02 & 48.97±1.16 & 46.19±0.0 & 48.88±0.0 & 48.7±0.68 & \textbf{49.33±0.0} \\
& & Wireless & \underline{93.8±0.45} & \textbf{94.0±0.24} & 93.47±0.61 & 93.27±0.55 & 90.8±1.04 & 92.2±0.51 & 92.0±0.0 & 93.33±0.0 & 93.2±0.3 & 90.33±0.0 \\
& & Abalone & 58.09±0.43 & 58.21±0.6 & 57.99±0.24 & 57.45±0.58 & 57.93±0.93 & 57.07±0.63 & 58.21±0.0 & \textbf{59.01±0.0} & \underline{58.28±0.6} & 57.74±0.0 \\
& & Wine quality & \textbf{52.57±1.05} & \underline{52.19±0.27} & 52.14±0.31 & 50.29±0.4 & 51.32±0.31 & 51.51±0.56 & 51.43±0.0 & 51.84±0.0 & 51.7±0.35 & 50.61±0.0 \\
& & Page blocks & 94.45±0.14 & \textbf{94.57±0.18} & 94.15±0.24 & 92.81±0.09 & 94.42±0.05 & 94.13±0.1 & 94.52±0.0 & 94.52±0.0 & \underline{94.57±0.33} & 94.28±0.0 \\
& & Electrical grid stability & \textbf{97.07±0.13} & \underline{97.01±0.13} & 96.43±0.06 & 91.75±0.58 & 94.8±0.0 & 95.35±0.1 & 96.07±0.0 & 96.0±0.0 & 95.85±0.46 & 94.8±0.0 \\
& & SUSY (small) & 75.48±0.07 & \underline{75.53±0.13} & 75.31±0.07 & -- & 74.2±0.18 & 74.97±0.18 & 74.81±0.0 & \textbf{75.59±0.0} & 75.37±0.11 & 74.55±0.0 \\
\cline{2-13}
& \multirow[c]{5}{*}{\rotatebox[origin=c]{90}{Categorical}} & Car & 69.36±0.97 & 69.62±0.38 & 69.87±1.11 & 70.0±0.0 & \textbf{70.9±1.24} & 70.0±0.0 & \underline{70.38±0.0} & \underline{70.38±0.0} & 70.38±0.38 & 69.62±0.0 \\
& & Phishing website & \textbf{83.91±0.28} & 82.76±0.49 & 77.67±1.03 & \underline{82.76±0.0} & 81.94±1.03 & 82.76±0.49 & 82.27±0.0 & 79.8±0.0 & 81.44±1.5 & 75.86±0.0 \\
& & Letter & 47.43±1.1 & 47.26±0.51 & 43.77±0.32 & 40.57±0.06 & 42.61±0.51 & 43.19±0.22 & \textbf{51.8±0.0} & \underline{47.77±0.0} & 47.43±1.08 & 43.17±0.0 \\
& & Chess & 25.99±0.33 & 26.33±0.21 & 25.75±0.49 & -- & \textbf{26.88±0.03} & 25.42±1.69 & 26.42±0.0 & \underline{26.51±0.0} & 26.17±0.83 & 24.04±0.0 \\
& & Connect & 65.88±0.01 & 65.9±0.06 & 65.95±0.04 & -- & 66.02±0.03 & 66.0±0.02 & 66.0±0.0 & \textbf{66.05±0.0} & \underline{66.04±0.03} & 65.89±0.0 \\
\cline{2-13}
& \multirow[c]{3}{*}{\rotatebox[origin=c]{90}{Mixed}} & Anuran & \underline{91.7±0.35} & \textbf{91.73±0.19} & 90.74±0.24 & 89.26±0.83 & 90.15±0.14 & 86.64±0.37 & 91.11±0.0 & 89.54±0.0 & 91.42±0.21 & 85.83±0.0 \\
& & Adult & \underline{81.58±0.63} & 81.47±0.32 & \textbf{82.01±0.19} & -- & 79.3±0.39 & 79.63±0.01 & 79.82±0.0 & 81.06±0.0 & 80.41±0.49 & 80.02±0.0 \\
& & Default credit card & 80.7±0.13 & \underline{80.73±0.19} & 80.65±0.05 & -- & 80.58±0.08 & 80.51±0.07 & \textbf{80.82±0.0} & 80.49±0.0 & 80.53±0.06 & 80.47±0.0 \\
\hline
\multirow[c]{15}{*}{\rotatebox[origin=c]{90}{\textbf{MAR}}} & \multirow[c]{7}{*}{\rotatebox[origin=c]{90}{Numerical}} & Yeast & 48.97±0.74 & 49.51±0.68 & 49.33±1.0 & \underline{50.49±0.68} & 48.34±2.14 & 50.13±1.4 & 50.22±0.0 & 48.43±0.0 & 49.33±1.93 & \textbf{50.67±0.0} \\
& & Wireless & \textbf{95.6±0.37} & 94.87±0.18 & 94.93±0.64 & 94.33±1.33 & 92.07±2.25 & 94.67±0.82 & 94.33±0.0 & 95.0±0.0 & \underline{95.27±0.37} & 92.33±0.0 \\
& & Abalone & \underline{58.66±0.57} & 58.37±0.37 & 58.18±0.44 & 57.26±0.52 & 58.15±0.79 & 57.58±0.44 & \textbf{60.13±0.0} & 58.37±0.0 & 58.56±0.73 & 56.78±0.0 \\
& & Wine quality & \underline{52.79±0.54} & 52.6±0.61 & 52.76±0.31 & 49.55±0.47 & 50.1±0.56 & 51.51±0.34 & \textbf{52.79±0.0} & 51.7±0.0 & 52.35±0.06 & 51.29±0.0 \\
& & Page blocks & 94.52±0.19 & \textbf{94.74±0.16} & 94.45±0.11 & 93.54±0.17 & 94.47±0.33 & 94.2±0.07 & 94.64±0.0 & 94.52±0.0 & \textbf{94.74±0.1} & 93.91±0.0 \\
& & Electrical grid stability & \underline{97.49±0.22} & \textbf{97.56±0.12} & 96.84±0.18 & 93.59±0.48 & 94.8±0.0 & 95.36±0.1 & 96.47±0.0 & 96.53±0.0 & 96.08±0.23 & 94.8±0.0 \\
& & SUSY (small) & 75.64±0.07 & 75.58±0.08 & 75.46±0.12 & -- & 75.17±0.18 & 75.31±0.1 & 75.24±0.0 & \underline{75.79±0.0} & 75.6±0.11 & \textbf 75.97±0.0 \\
\cline{2-13}
& \multirow[c]{5}{*}{\rotatebox[origin=c]{90}{Categorical}} & Car & 69.46±0.64 & 70.0±0.72 & 69.85±0.34 & \textbf{70.77±0.0} & 70.08±0.63 & 70.0±0.0 & \textbf{70.77±0.0} & 70.38±0.0 & 70.54±0.34 & 68.46±0.0 \\
& & Phishing website & 83.25±0.49 & \underline{83.58±0.28} & 80.79±1.71 & 82.27±0.0 & 81.28±0.85 & 81.28±0.0 & \textbf{85.71±0.0} & 80.3±0.0 & 82.76±1.3 & 77.83±0.0 \\
& & Letter & 48.77±1.05 & \underline{49.31±0.41} & 46.12±1.17 & 42.41±0.02 & 38.0±0.71 & 45.12±0.43 & \textbf{52.13±0.0} & 47.97±0.0 & 46.61±0.91 & 45.07±0.0 \\
& & Chess & 26.46±0.59 & 26.05±0.84 & 25.89±0.85 & -- & 27.04±0.06 & 25.02±0.75 & 26.94±0.0 & \textbf{27.25±0.0} & \underline{27.13±0.51} & 23.78±0.0 \\
& & Connect & 66.05±0.03 & 66.05±0.05 & 66.0±0.02 & -- & 66.0±0.06 & 66.0±0.01 & \textbf{66.09±0.0} & \underline{66.08±0.0} & 66.07±0.02 & 65.92±0.0 \\
\cline{2-13}
& \multirow[c]{3}{*}{\rotatebox[origin=c]{90}{Mixed}} & Anuran & 91.82±0.23 & \underline{91.91±0.77} & 90.96±0.23 & 89.69±0.27 & 88.8±0.24 & 86.94±0.19 & 91.11±0.0 & 89.54±0.0 & \textbf{91.94±0.19} & 87.78±0.0 \\
& & Adult & 81.27±0.21 & \underline{81.36±0.5} & \textbf{81.67±0.17} & -- & 79.1±0.81 & 79.64±0.04 & 80.91±0.0 & 80.32±0.0 & 81.22±0.27 & 80.13±0.0 \\
& & Default credit card & 80.64±0.14 & 80.61±0.09 & \textbf{80.81±0.08} & -- & 80.64±0.11 & \underline{80.67±0.07} & 80.64±0.0 & \underline{80.67±0.0} & 80.56±0.04 & 80.6±0.0 \\
\hline
\end{tabular}
\label{table:RF_accuracy}
\end{adjustbox}
%}
\end{center}
\end{table}

\begin{table}[h]
\begin{center}
\caption{Imputation performance in terms of RMSE (Average±Std, averaged over 5 runs) under MCAR, MNAR and MAR assumptions (the case in which $20\%$ of entries are missing). The best results for each dataset are highlighted in bold font, and the second-best result is underlined.}
\begin{adjustbox}{width=\textwidth}
\begin{tabular}{|c|c|c| c c c c c c c c c c| }%# *{13}{c}
\hline
{\textbf{Miss.}} & \textbf{Data type}& {\textbf{Dataset}} & {\textbf{EGG-GAE}} & {\textbf{k-EGG-GAE}} & {\textbf{NN}} & {\textbf{GINN}} & {\textbf{GAIN}} & {\textbf{MIDA}} & {\textbf{KNNI}} & {\textbf{MICE}} & {\textbf{MF}} & {\textbf{Mean}} \\
\hline
\multirow[c]{10}{*}{\rotatebox[origin=c]{90} {\textbf{MCAR}}} & \multirow[c]{7}{*}{\rotatebox[origin=c]{90} {\textbf{Numerical}}} & Yeast & 0.9343±0.0069 & \underline{0.9271±0.0099} & 0.9431±0.0063 & 1.0749±0.0401 & 1.061±0.0052 & 0.9825±0.0031 & 1.014±0.0 & 0.9315±0.0 & \textbf{0.9222±0.0259} & 0.9987±0.0 \\
& & Wireless & \underline{0.6082±0.001} & \textbf{0.6067±0.0102} & 0.645±0.0091 & 1.078±0.007 & 0.8865±0.1909 & 0.8097±0.0176 & 0.7341±0.0 & 0.6351±0.0 & 0.6425±0.0156 & 0.9851±0.0 \\
& & Abalone & 0.3982±0.0036 & \underline{0.3925±0.0025} & 0.4586±0.0046 & 1.5066±0.2678 & 0.6314±0.0608 & 0.5183±0.0018 & 0.4931±0.0 & 0.4051±0.0 & \textbf{0.3747±0.0063} & 0.9781±0.0 \\
& & Wine quality & \underline{0.803±0.0047} & \textbf{0.8006±0.0032} & 0.8398±0.011 & 1.3678±0.055 & 0.9491±0.0066 & 0.9125±0.0065 & 0.8658±0.0 & 0.8268±0.0 & 0.8204±0.0034 & 1.0314±0.0 \\
& & Page blocks & \underline{0.6211±0.0048} & 0.6218±0.0056 & 0.6673±0.0084 & 1.1888±0.0113 & 0.9939±0.0047 & 0.8385±0.0059 & 0.6813±0.0 & 0.7527±0.0 & \textbf{0.6056±0.0273} & 1.0947±0.0 \\
& & Electrical grid stability & \textbf{0.8587±0.0013} & \underline{0.8588±0.0028} & 0.8862±0.0006 & 1.2907±0.0247 & 0.9902±0.0253 & 0.9768±0.0026 & 0.9966±0.0 & 0.9018±0.0 & 0.9112±0.0047 & 1.0193±0.0 \\
& & SUSY (small) & \underline{0.5781±0.0014} & \textbf{0.5772±0.0038} & 0.6542±0.003 & -- & 0.881±0.0468 & 0.777±0.0029 & 0.7078±0.0 & 0.666±0.0 & 0.638±0.0019 & 1.0195±0.0 \\

\cline{2-13}
& \multirow[c]{3}{*}{\rotatebox[origin=c]{90} {\textbf{Mixed}}} & Anuran & 0.4978±0.0104 & \underline{0.4966±0.0037} & 0.5878±0.0032 & 1.0423±0.0112 & 0.8446±0.0169 & 0.9867±0.0275 & \textbf{0.4808±0.0} & 0.5893±0.0 & 0.5431±0.0087 & 1.0405±0.0 \\
& & Adult & \underline{0.8568±0.0027} & \textbf{0.8538±0.0032} & 0.8746±0.0004 & -- & 0.9983±0.0063 & 0.9928±0.0014 & 0.9321±0.0 & 0.9574±0.0 & 0.9132±0.0065 & 0.9999±0.0 \\
& & Default credit card & \underline{0.6309±0.0222} & \textbf{0.6224±0.0116} & 0.7086±0.0031 & -- & 1.0293±0.2887 & 0.9621±0.01 & 0.7338±0.0 & 0.6802±0.0 & 0.641±0.0142 & 1.0121±0.0 \\
\hline
\multirow[c]{10}{*}{\rotatebox[origin=c]{90} {\textbf{MNAR}}} & \multirow[c]{7}{*}{\rotatebox[origin=c]{90} {\textbf{Numerical}}} & Yeast & 0.7787±0.0087 & 0.7739±0.0083 & \underline{0.7739±0.0006} & 0.9819±0.0332 & 0.9562±0.0206 & 0.7934±0.0022 & 0.8633±0.0 & \textbf{0.7396±0.0} & 0.875±0.0282 & 0.8295±0.0 \\
& & Wireless & \underline{0.6864±0.008} & \textbf{0.6806±0.0023} & 0.717±0.0061 & 1.2162±0.0199 & 0.9905±0.1485 & 0.8836±0.0274 & 0.7612±0.0 & 0.7016±0.0 & 0.7213±0.0102 & 1.0959±0.0 \\
& & Abalone & 0.4032±0.0045 & 0.4072±0.0086 & 0.5075±0.0095 & 1.3258±0.0264 & 0.6234±0.0682 & 0.5521±0.0012 & 0.4503±0.0 & \underline{0.3854±0.0} & \textbf{0.3836±0.0049} & 1.116±0.0 \\
& & Wine quality & \underline{0.7693±0.0161} & \textbf{0.7581±0.0051} & 0.7761±0.0034 & 1.1364±0.0258 & 0.9264±0.0017 & 0.853±0.0055 & 0.8137±0.0 & 0.8722±0.0 & 0.7786±0.0071 & 0.9738±0.0 \\
& & Page blocks & \underline{0.6971±0.0149} & \textbf{0.6842±0.0053} & 0.7284±0.003 & 1.373±0.0246 & 1.3086±0.1741 & 0.9244±0.0056 & 0.7626±0.0 & 0.8433±0.0 & 0.7396±0.0855 & 1.1922±0.0 \\
& & Electrical grid stability & \underline{0.8556±0.0017} & \textbf{0.8543±0.001} & 0.8835±0.0007 & 1.403±0.0207 & 1.0091±0.0366 & 0.9646±0.0048 & 1.0049±0.0 & 0.8952±0.0 & 0.9196±0.006 & 1.0114±0.0 \\
& & SUSY (small) & \underline{0.5785±0.006} & \textbf{0.5752±0.0028} & 0.6589±0.0026 & -- & 1.0478±0.0462 & 0.7746±0.0031 & 0.7022±0.0 & 0.6726±0.0 & 0.6357±0.0042 & 1.0178±0.0 \\

\cline{2-13}
& \multirow[c]{3}{*}{\rotatebox[origin=c]{90} {\textbf{Mixed}}} & Anuran & \underline{0.4274±0.0081} & 0.4335±0.0052 & 0.5214±0.003 & 1.0255±0.0124 & 0.8864±0.0626 & 1.0009±0.0098 & \textbf{0.4092±0.0} & 0.5027±0.0 & 0.4878±0.0048 & 1.0541±0.0 \\
& & Adult & \underline{0.8643±0.0057} & \textbf{0.8639±0.0043} & 0.883±0.0019 & -- & 1.0271±0.0023 & 1.0036±0.0003 & 0.9693±0.0 & 0.9772±0.0 & 0.9665±0.0379 & 1.0107±0.0 \\
& & Default credit card & \textbf{0.6949±0.0038} & \underline{0.6953±0.001} & 0.7409±0.0016 & -- & 1.366±0.1709 & 1.0112±0.0114 & 0.7515±0.0 & 0.8555±0.0 & 0.7063±0.0159 & 1.0731±0.0 \\
\hline
\multirow[c]{10}{*}{\rotatebox[origin=c]{90} {\textbf{MAR}}} & \multirow[c]{7}{*}{\rotatebox[origin=c]{90} {\textbf{Numerical}}} & Yeast & 0.8196±0.019 & 0.8163±0.0128 & \underline{0.8059±0.0057} & 0.8575±0.0178 & 1.0849±0.0917 & 0.8085±0.0019 & 0.9018±0.0 & \textbf{0.7909±0.0} & 0.8748±0.0346 & 0.8503±0.0 \\
& & Wireless & \underline{0.6408±0.0104} & \textbf{0.635±0.0026} & 0.6805±0.0036 & 1.2775±0.0159 & 1.0775±0.1336 & 0.8625±0.0203 & 0.644±0.0 & 0.6748±0.0 & 0.6772±0.0153 & 1.125±0.0 \\
& & Abalone & 0.3913±0.0038 & 0.4003±0.0057 & 0.4984±0.0124 & 1.3368±0.028 & 0.7047±0.1387 & 0.5436±0.0043 & 0.4088±0.0 & \textbf{0.3837±0.0} & \underline{0.3851±0.013} & 1.1543±0.0 \\
& & Wine quality & 0.7018±0.0073 & \textbf{0.6949±0.0034} & 0.7079±0.0033 & 1.1996±0.0635 & 1.2254±0.0999 & 0.8075±0.0065 & 0.7125±0.0 & 0.7295±0.0 & \underline{0.6994±0.0018} & 0.9547±0.0 \\
& & Page blocks & 0.68±0.0037 & \underline{0.6718±0.0105} & 0.7535±0.0082 & 1.386±0.0058 & 1.6437±0.7462 & 0.9527±0.0069 & \textbf{0.667±0.0} & 0.8807±0.0 & 0.6744±0.0449 & 1.2318±0.0 \\
& & Electrical grid stability & \underline{0.8141±0.0009} & \textbf{0.813±0.0026} & 0.8483±0.0007 & 1.3966±0.0251 & 1.0399±0.0344 & 0.9533±0.0069 & 0.96±0.0 & 0.8318±0.0 & 0.8875±0.0021 & 1.0124±0.0 \\
& & SUSY (small) & \textbf{0.5213±0.0026} & \underline{0.5222±0.0023} & 0.6206±0.005 & -- & 1.1357±0.0857 & 0.7572±0.0033 & 0.6158±0.0 & 0.6116±0.0 & 0.602±0.0016 & 1.0217±0.0 \\

\cline{2-13}
& \multirow[c]{3}{*}{\rotatebox[origin=c]{90} {\textbf{Mixed}}} & Anuran & \underline{0.4363±0.0023} & 0.4401±0.006 & 0.5353±0.0048 & 1.0484±0.0188 & 1.7316±0.1103 & 1.0338±0.0154 & \textbf{0.4201±0.0} & 0.5038±0.0 & 0.514±0.0012 & 1.0968±0.0 \\
& & Adult & \underline{0.8333±0.0027} & \textbf{0.8323±0.0017} & 0.8672±0.0015 & -- & 1.514±0.5263 & 1.0078±0.0027 & 0.8603±0.0 & 0.9779±0.0 & 0.8501±0.0059 & 1.012±0.0 \\
& & Default credit card & 0.7012±0.0015 & \underline{0.6983±0.0065} & 0.7514±0.0033 & -- & 1.2989±0.2358 & 0.9903±0.0306 & 0.7488±0.0 & 0.7081±0.0 & \textbf{0.6974±0.025} & 1.0719±0.0 \\
\hline
\end{tabular}
\label{table:imp_rmse}
\end{adjustbox}
\end{center}
\end{table}
\begin{table}[h]
\begin{center}
\caption{Imputation performance in terms of MAE (Average±Std, averaged over 5 runs) under MCAR, MNAR and MAR assumptions (the case in which $20\%$ of entries are missing). The best results for each dataset are highlighted in bold font, and the second-best result is underlined.}
\begin{adjustbox}{width=\textwidth}
\begin{tabular}{|c|c|c| c c c c c c c c c c| }%# *{13}{c}
\hline
{\textbf{Miss.}} & \textbf{Data type} & {\textbf{Dataset}} & {\textbf{EGG-GAE}} & {\textbf{k-EGG-GAE}} & {\textbf{NN}} & {\textbf{GINN}} & {\textbf{GAIN}} & {\textbf{MIDA}} & {\textbf{KNNI}} & {\textbf{MICE}} & {\textbf{MF}} & {\textbf{Mean}} \\
\hline
\multirow[c]{10}{*}{\rotatebox[origin=c]{90} {\textbf{MCAR}}} & \multirow[c]{7}{*}{\rotatebox[origin=c]{90} {\textbf{Numerical}}} & Yeast & 0.5299±0.0071 & \underline{0.5256±0.0104} & 0.5459±0.0083 & 0.5957±0.0335 & 0.6128±0.0557 & 0.5587±0.0038 & 0.5728±0.0 & 0.5258±0.0 & 0.5279±0.0098 & \textbf 0.5217±0.0 \\
& &  Wireless & \underline{0.4469±0.0022} & \textbf{0.445±0.0064} & 0.4884±0.0071 & 0.8302±0.0046 & 0.7233±0.1179 & 0.6142±0.0213 & 0.5467±0.0 & 0.4737±0.0 & 0.4593±0.0116 & 0.7771±0.0 \\
& & Abalone & 0.2673±0.0064 & 0.2577±0.002 & 0.3396±0.0071 & 1.0507±0.1161 & 0.563±0.0938 & 0.3687±0.0013 & 0.3004±0.0 & \underline{0.2414±0.0} & \textbf{0.2306±0.0013} & 0.7755±0.0 \\
& & Wine quality & \textbf{0.5596±0.0091} & \underline{0.5597±0.005} & 0.5849±0.0076 & 1.086±0.0488 & 0.6939±0.0046 & 0.6476±0.0067 & 0.6014±0.0 & 0.5699±0.0 & 0.565±0.0027 & 0.7499±0.0 \\
& & Page blocks & 0.3204±0.0064 & 0.3181±0.0071 & 0.3675±0.0074 & 0.7765±0.01 & 0.6148±0.0043 & 0.4683±0.0032 & \underline{0.2902±0.0} & 0.4091±0.0 & \textbf{0.2711±0.009} & 0.605±0.0 \\
& & Electrical grid stability & 0.7148±0.0011 & \underline{0.713±0.0033} & 0.7444±0.0004 & 1.059±0.0175 & 0.8501±0.0276 & 0.8328±0.0008 & 0.8297±0.0 & \textbf{0.6791±0.0} & 0.7427±0.0042 & 0.8809±0.0 \\
& & SUSY (small) & \textbf{0.3776±0.0014} & \underline{0.3798±0.005} & 0.4484±0.003 & -- & 0.6804±0.0275 & 0.5487±0.0032 & 0.4436±0.0 & 0.4469±0.0 & 0.4152±0.0015 & 0.7453±0.0 \\

\cline{2-13}
& \multirow[c]{3}{*}{\rotatebox[origin=c]{90} {\textbf{Mixed}}} &  Anuran & 0.2905±0.0036 & \underline{0.2898±0.0041} & 0.3643±0.0026 & 0.7337±0.0097 & 0.6273±0.0079 & 0.7068±0.0383 & \textbf{0.2418±0.0} & 0.3317±0.0 & 0.3173±0.0024 & 0.7442±0.0 \\
& & Adult & \textbf{0.4774±0.0045} & \underline{0.4775±0.0058} & 0.5091±0.0033 & -- & 0.6029±0.0048 & 0.5867±0.0022 & 0.5344±0.0 & 0.6024±0.0 & 0.4919±0.005 & 0.5391±0.0 \\
& & Default credit card & 0.2597±0.0136 & 0.2559±0.0101 & 0.3104±0.0028 & -- & 0.4818±0.0399 & 0.5039±0.0154 & \underline{0.2345±0.0} & 0.2505±0.0 & \textbf{0.209±0.002} & 0.4585±0.0 \\
\hline
\multirow[c]{10}{*}{\rotatebox[origin=c]{90} {\textbf{MNAR}}} & \multirow[c]{7}{*}{\rotatebox[origin=c]{90} {\textbf{Numerical}}} & Yeast & 0.5701±0.013 & 0.5624±0.0079 & 0.5587±0.005 & 0.6436±0.0216 & 0.6682±0.0521 & 0.5415±0.0028 & 0.5794±0.0 & \underline{0.5305±0.0} & 0.5891±0.0159 & \textbf 0.5261±0.0 \\
& &  Wireless & 0.5088±0.0074 & \textbf{0.5052±0.0026} & 0.5384±0.0061 & 0.9311±0.0199 & 0.8447±0.095 & 0.6793±0.0225 & 0.5678±0.0 & 0.5354±0.0 & \underline{0.5075±0.0085} & 0.8922±0.0 \\
& & Abalone & 0.2797±0.0043 & 0.2891±0.0075 & 0.3657±0.0039 & 1.0653±0.0251 & 0.6671±0.076 & 0.3918±0.0019 & 0.2913±0.0 & \textbf{0.2331±0.0} & \underline{0.2426±0.0042} & 0.8826±0.0 \\
& & Wine quality & 0.5534±0.0055 & \textbf{0.5521±0.0117} & 0.5678±0.0038 & 0.8638±0.0222 & 0.7241±0.0198 & 0.6385±0.0039 & 0.5883±0.0 & 0.6033±0.0 & \textbf{0.5521±0.0039} & 0.7288±0.0 \\
& & Page blocks & 0.3463±0.0051 & 0.3422±0.0072 & 0.3912±0.0049 & 0.8623±0.0159 & 0.6402±0.0183 & 0.5015±0.0049 & \underline{0.3353±0.0} & 0.4598±0.0 & \textbf{0.292±0.0108} & 0.675±0.0 \\
& & Electrical grid stability & 0.7045±0.002 & \underline{0.7028±0.0009} & 0.7384±0.0011 & 1.1405±0.0153 & 0.8499±0.0252 & 0.8244±0.0053 & 0.8282±0.0 & \textbf{0.6789±0.0} & 0.7474±0.005 & 0.8736±0.0 \\
& & SUSY (small) & \underline{0.3739±0.0028} & \textbf{0.3734±0.0011} & 0.4472±0.0023 & -- & 0.7795±0.0224 & 0.5495±0.0037 & 0.4404±0.0 & 0.4475±0.0 & 0.4102±0.002 & 0.7438±0.0 \\

\cline{2-13}
& \multirow[c]{3}{*}{\rotatebox[origin=c]{90} {\textbf{Mixed}}} & Anuran & \underline{0.2876±0.0059} & 0.2899±0.006 & 0.3628±0.0018 & 0.7899±0.0023 & 0.6611±0.0784 & 0.7679±0.01 & \textbf{0.2501±0.0} & 0.3324±0.0 & 0.3172±0.0025 & 0.7911±0.0 \\
& & Adult & \underline{0.4815±0.0071} & \textbf{0.4797±0.01} & 0.5055±0.001 & -- & 0.6669±0.0674 & 0.5942±0.0071 & 0.5597±0.0 & 0.6143±0.0 & 0.5213±0.0136 & 0.5427±0.0 \\
& & Default credit card & 0.2538±0.0058 & 0.2575±0.0008 & 0.2916±0.0036 & -- & 0.6308±0.0385 & 0.5052±0.0148 & \underline{0.2308±0.0} & 0.3219±0.0 & \textbf{0.2183±0.0033} & 0.47±0.0 \\
\hline
\multirow[c]{10}{*}{\rotatebox[origin=c]{90} {\textbf{MAR}}} & \multirow[c]{7}{*}{\rotatebox[origin=c]{90} {\textbf{Numerical}}} & Yeast & 0.6059±0.0173 & 0.604±0.0119 & 0.5829±0.01 & 0.5925±0.0175 & 1.0058±0.2793 & \underline{0.5673±0.004} & 0.6328±0.0 & 0.5699±0.0 & 0.6061±0.0165 & \textbf 0.5566±0.0 \\
& &  Wireless & 0.4746±0.0074 & \underline{0.4743±0.0035} & 0.511±0.0037 & 0.9639±0.0139 & 0.929±0.0459 & 0.6702±0.0171 & 0.4759±0.0 & 0.5158±0.0 & \textbf{0.4711±0.0091} & 0.9157±0.0 \\
& & Abalone & 0.2921±0.0132 & 0.3001±0.0069 & 0.3711±0.0066 & 1.0879±0.0348 & 0.652±0.0891 & 0.3973±0.004 & 0.2709±0.0 & \textbf{0.2409±0.0} & \underline{0.2514±0.0039} & 0.9208±0.0 \\
& & Wine quality & 0.5113±0.0096 & \underline{0.4977±0.0075} & 0.5208±0.0022 & 0.9236±0.0389 & 0.9259±0.0472 & 0.6061±0.0076 & 0.5052±0.0 & 0.5137±0.0 & \textbf{0.497±0.0059} & 0.7183±0.0 \\
& & Page blocks & 0.3036±0.0092 & 0.2947±0.0068 & 0.3459±0.0025 & 0.8488±0.0044 & 0.8011±0.1234 & 0.4557±0.0075 & \textbf{0.2212±0.0} & 0.422±0.0 & \underline{0.2519±0.013} & 0.639±0.0 \\
& & Electrical grid stability & 0.6574±0.0012 & \underline{0.6568±0.0019} & 0.7018±0.0023 & 1.1384±0.0174 & 0.898±0.0155 & 0.8115±0.0053 & 0.7875±0.0 & \textbf{0.6158±0.0} & 0.7153±0.0082 & 0.873±0.0 \\
& & SUSY (small) & \underline{0.3274±0.0014} & \textbf{0.3243±0.0006} & 0.4095±0.0049 & -- & 0.8485±0.0269 & 0.5244±0.0028 & 0.3751±0.0 & 0.3941±0.0 & 0.3793±0.0021 & 0.7507±0.0 \\

\cline{2-13}
& \multirow[c]{3}{*}{\rotatebox[origin=c]{90} {\textbf{Mixed}}} & Anuran & 0.2938±0.0034 & \underline{0.293±0.0024} & 0.3707±0.0033 & 0.8179±0.0084 & 1.3936±0.0946 & 0.8071±0.0185 & \textbf{0.2637±0.0} & 0.3414±0.0 & 0.3409±0.0032 & 0.8439±0.0 \\
& & Adult & \underline{0.4113±0.0096} & 0.4122±0.0104 & 0.4616±0.001 & -- & 0.7908±0.2128 & 0.5886±0.0071 & 0.4591±0.0 & 0.6054±0.0 & \textbf{0.3816±0.003} & 0.5429±0.0 \\
& & Default credit card & 0.2301±0.0097 & 0.225±0.0071 & 0.2649±0.0013 & -- & 0.8806±0.0318 & 0.4543±0.0384 & \underline{0.1925±0.0} & 0.2108±0.0 & \textbf{0.1794±0.0029} & 0.4238±0.0 \\
\hline
\end{tabular}
\label{table:imp_mae}
\end{adjustbox}
\end{center}
\end{table}

%
% Imputation Cat
\begin{table}[h]
\begin{center}
\caption{Imputation performance in terms of Accuracy of reconstructed categorical variables (Average±Std, averaged over 5 runs) under MCAR, MNAR and MAR assumptions (the case in which $20\%$ of entries are missing). The best results for each dataset are highlighted in bold font, and the second-best result is underlined.}
\begin{adjustbox}{width=\textwidth}
\begin{tabular}{|c|c|c| c c c c c c c c c c| }%# *{13}{c}
\hline
{\textbf{Miss.}} & \textbf{Dataset type}& {\textbf{Dataset}} & {\textbf{EGG-GAE}} & {\textbf{k-EGG-GAE}} & {\textbf{NN}} & {\textbf{GINN}} & {\textbf{GAIN}} & {\textbf{MIDA}} & {\textbf{KNNI}} & {\textbf{MICE}} & {\textbf{MF}} & {\textbf{Mean}} \\
\hline
\multirow[c]{8}{*}{\rotatebox[origin=c]{90}{\textbf{MCAR}}} & \multirow[c]{5}{*}{\rotatebox[origin=c]{90}{Categorical}} & car & 26.6±1.95 & 22.65±1.03 & 26.92±1.28 & 23.08±0.55 & 30.98±0.49 & \underline{32.05±1.47} & 26.28±0.0 & 27.24±0.0 & 29.28±1.03 & \textbf{33.33±0.0} \\
& & phishing website & \underline{60.64±1.56} & \textbf{61.19±0.57} & 57.63±3.47 & 51.69±0.88 & 46.21±4.19 & 41.82±1.11 & 48.49±0.0 & 49.59±0.0 & 49.13±2.8 & 49.04±0.0 \\
& & letter & \underline{41.1±0.61} & 40.88±0.35 & 30.12±0.73 & 18.2±0.01 & 22.98±0.95 & 24.23±0.68 & \textbf{45.75±0.0} & 26.97±0.0 & 31.49±0.57 & 24.52±0.0 \\
& & chess & 23.27±0.21 & \textbf{23.83±0.62} & 23.58±0.41 & 20.54±0.03 & 19.38±0.27 & 19.16±0.22 & 20.2±0.0 & 22.06±0.0 & 20.57±0.42 & \underline{23.68±0.0} \\
& & connect & \underline{88.86±0.13} & \textbf{88.9±0.05} & 86.67±0.14 & -- & 77.97±3.76 & 83.82±0.03 & 82.75±0.0 & 83.8±0.0 & 84.44±0.35 & 81.49±0.0 \\
\cline{2-13}
& \multirow[c]{3}{*}{\rotatebox[origin=c]{90}{Mixed}} &  anuran & 82.98±0.84 & \underline{83.23±0.48} & 68.73±2.0 & 40.36±0.0 & 31.37±0.35 & 23.64±12.18 & \textbf{83.28±0.0} & 40.96±0.0 & 67.47±5.39 & 41.27±0.0 \\
& & Adult & \textbf{72.64±0.54} & \underline{72.58±0.43} & 69.18±0.23 & -- & 32.61±2.41 & 34.54±1.81 & 42.36±0.0 & 19.64±0.0 & 33.39±0.31 & 55.43±0.0 \\
& & default credit card & \underline{67.82±0.18} & \textbf{68.08±0.37} & 65.92±0.19 & -- & 48.98±0.2 & 44.85±9.16 & 58.31±0.0 & 46.12±0.0 & 50.86±2.67 & 48.89±0.0 \\
\hline
\multirow[c]{8}{*}{\rotatebox[origin=c]{90}{\textbf{MNAR}}} & \multirow[c]{5}{*}{\rotatebox[origin=c]{90}{Categorical}} & car & 23.82±0.64 & 23.52±2.77 & 24.44±3.25 & \textbf{32.62±0.17} & \underline{30.16±0.94} & 29.45±0.0 & 25.46±0.0 & 29.75±0.0 & 26.58±2.09 & 28.83±0.0 \\
& & phishing website & \textbf{61.46±0.72} & \underline{60.8±0.49} & 55.3±1.85 & 52.65±0.16 & 45.74±3.12 & 42.33±0.75 & 48.01±0.0 & 46.31±0.0 & 46.59±1.24 & 53.98±0.0 \\
& & letter & \underline{41.67±0.45} & 41.03±0.33 & 30.85±0.48 & 18.18±0.01 & 22.33±1.92 & 24.18±0.55 & \textbf{46.01±0.0} & 26.51±0.0 & 32.2±0.58 & 24.95±0.0 \\
& & chess & \textbf{22.54±0.45} & \underline{22.35±0.19} & 22.2±0.72 & -- & 18.67±0.1 & 18.36±0.08 & 18.65±0.0 & 19.76±0.0 & 19.57±0.71 & 21.45±0.0 \\
& & connect & \underline{88.85±0.23} & \textbf{88.92±0.04} & 86.75±0.1 & -- & 80.55±4.75 & 84.1±0.03 & 83.09±0.0 & 84.05±0.0 & 83.91±0.06 & 81.86±0.0 \\
\cline{2-13}
& \multirow[c]{3}{*}{\rotatebox[origin=c]{90}{Mixed}} & anuran & 84.01±0.77 & \underline{84.74±0.32} & 71.41±1.53 & 62.66±0.0 & 28.91±2.56 & 15.31±7.17 & \textbf{85.62±0.0} & 43.44±0.0 & 71.62±0.24 & 45.78±0.0 \\
& &  Adult & \underline{73.76±0.53} & \textbf{73.95±0.43} & 70.8±0.19 & -- & 32.5±3.64 & 37.14±1.11 & 36.47±0.0 & 22.9±0.0 & 36.58±1.53 & 58.14±0.0 \\
& & default credit card & \textbf{66.79±0.35} & \underline{66.38±0.15} & 64.9±0.26 & -- & 42.42±1.49 & 44.54±9.2 & 56.47±0.0 & 42.78±0.0 & 52.02±2.62 & 48.51±0.0 \\
\hline
\multirow[c]{8}{*}{\rotatebox[origin=c]{90}{\textbf{MAR}}} & \multirow[c]{5}{*}{\rotatebox[origin=c]{90}{Categorical}} & car & 23.9±2.05 & 23.61±2.37 & 24.33±3.62 & \textbf{33.07±0.32} & \underline{31.77±1.6} & 28.88±0.0 & 25.27±0.0 & 29.6±0.0 & 28.45±3.1 & 29.24±0.0 \\
& & phishing website & \textbf{63.47±0.94} & \underline{62.88±1.95} & 60.05±1.95 & 53.07±0.2 & 45.86±0.73 & 36.17±0.0 & 51.77±0.0 & 45.74±0.0 & 43.03±1.14 & 54.61±0.0 \\
& & letter & \underline{44.03±0.63} & 43.69±0.05 & 32.65±0.65 & 20.95±0.01 & 18.96±0.8 & 24.38±0.39 & \textbf{52.2±0.0} & 28.19±0.0 & 34.99±0.43 & 26.68±0.0 \\
& & chess & \textbf{17.84±0.55} & \underline{17.62±0.09} & 17.57±0.59 & -- & 16.59±0.21 & 16.23±0.39 & 15.71±0.0 & 16.67±0.0 & 16.14±0.49 & 17.26±0.0 \\
& & connect & \textbf{91.58±0.27} & \underline{91.57±0.23} & 89.24±0.16 & -- & 83.63±3.84 & 87.2±0.06 & 86.38±0.0 & 87.25±0.0 & 87.87±0.13 & 84.45±0.0 \\
\cline{2-13}
& \multirow[c]{3}{*}{\rotatebox[origin=c]{90}{Mixed}} & anuran & 77.51±0.36 & \underline{78.62±1.33} & 62.0±2.03 & 54.0±0.14 & 34.68±2.11 & 6.33±0.9 & \textbf{80.76±0.0} & 30.64±0.0 & 59.86±1.33 & 41.81±0.0 \\
& &  Adult & \textbf{73.83±0.62} & \underline{73.13±0.34} & 69.57±0.13 & -- & 21.24±7.07 & 29.84±2.02 & 38.97±0.0 & 20.67±0.0 & 37.34±1.76 & 56.69±0.0 \\
& & default credit card & \underline{62.44±0.08} & \textbf{62.48±0.33} & 60.97±0.41 & -- & 36.67±2.93 & 43.98±8.64 & 53.97±0.0 & 43.91±0.0 & 51.22±2.8 & 46.41±0.0 \\
\hline
\end{tabular}
\label{table:imp_catacc}
\end{adjustbox}
\end{center}
\end{table}

\section{ASSUMPTIONS AND LIMITATIONS}
\label{sec:assumptions_and_limitations}

1) The pairwise calculation in the sampling procedure of the EGG block requires $n^2$ operations, so increasing the batch size results in a quadratic increase in training/inference time. In addition, we believe that the procedure could generally be replaced with a learnable block. In fact, we believe that the sampling process should be iterable, such that the first iteration provides an initial approximation of the neighbourhood and subsequent iterations eliminate noisy neighbours.

2) In the paper we rely on a pretty simple graph construction approach: from the sampled batch $\mathrm{X}$ we construct for each node its neighbourhood and pass the obtained graph through a GNN head. There are a number of modifications that will allow to obtain better gradients, resulting in a better solution. For example, we can extract $\mathrm{b}$ combinations of $\mathrm{m}$ rows from the obtained matrix $\mathrm{P}$ (with/without repetitions), and then carry out the subsequent operations described in Section \ref{sec:method} without modification. Such modification will allows us to obtain multiple predictions for the same data point in a single pass, resulting in a theoretically better gradient.
\clearpage
\bibliographystyle{apalike}
\bibliography{biblio.bib}

\begin{thebibliography}{}

\bibitem[Acuna and Rodriguez, 2004]{acuna2004treatment}
Acuna, E. and Rodriguez, C. (2004).
\newblock The treatment of missing values and its effect on classifier
  accuracy.
\newblock In {\em Classification, clustering, and data mining applications},
  pages 639--647. Springer.

\bibitem[Azur et~al., 2011]{azur2011multiple}
Azur, M.~J., Stuart, E.~A., Frangakis, C., and Leaf, P.~J. (2011).
\newblock Multiple imputation by chained equations: what is it and how does it
  work?
\newblock {\em International journal of methods in psychiatric research},
  20(1):40--49.

\bibitem[Bengio and Gingras, 1995]{bengio1995recurrent}
Bengio, Y. and Gingras, F. (1995).
\newblock Recurrent neural networks for missing or asynchronous data.
\newblock {\em Advances in neural information processing systems}, 8.

\bibitem[Bertsimas et~al., 2017]{bertsimas2017predictive}
Bertsimas, D., Pawlowski, C., and Zhuo, Y.~D. (2017).
\newblock From predictive methods to missing data imputation: an optimization
  approach.
\newblock {\em J. Mach. Learn. Res.}, 18(1):7133--7171.

\bibitem[Bianchi et~al., 2021]{bianchi2021graph}
Bianchi, F.~M., Grattarola, D., Livi, L., and Alippi, C. (2021).
\newblock Graph neural networks with convolutional arma filters.
\newblock {\em IEEE Transactions on Pattern Analysis and Machine Intelligence}.

\bibitem[Breiman, 2001]{breiman2001random}
Breiman, L. (2001).
\newblock Random forests.
\newblock {\em Machine learning}, 45(1):5--32.

\bibitem[Chen et~al., 2020]{chen2020learning}
Chen, X., Chen, S., Yao, J., Zheng, H., Zhang, Y., and Tsang, I.~W. (2020).
\newblock Learning on attribute-missing graphs.
\newblock {\em IEEE transactions on pattern analysis and machine intelligence}.

\bibitem[Cong et~al., 2020]{cong2020minimal}
Cong, W., Forsati, R., Kandemir, M., and Mahdavi, M. (2020).
\newblock Minimal variance sampling with provable guarantees for fast training
  of graph neural networks.
\newblock In {\em 26th ACM SIGKDD International Conference on Knowledge
  Discovery \& Data Mining}, pages 1393--1403.

\bibitem[Cosmo et~al., 2020]{cosmo2020latent}
Cosmo, L., Kazi, A., Ahmadi, S.-A., Navab, N., and Bronstein, M. (2020).
\newblock Latent-graph learning for disease prediction.
\newblock In {\em International Conference on Medical Image Computing and
  Computer-Assisted Intervention}, pages 643--653. Springer.

\bibitem[Farhangfar et~al., 2007]{farhangfar2007novel}
Farhangfar, A., Kurgan, L.~A., and Pedrycz, W. (2007).
\newblock A novel framework for imputation of missing values in databases.
\newblock {\em IEEE Transactions on Systems, Man, and Cybernetics-Part A:
  Systems and Humans}, 37(5):692--709.

\bibitem[Gilmer et~al., 2017]{gilmer2017neural}
Gilmer, J., Schoenholz, S.~S., Riley, P.~F., Vinyals, O., and Dahl, G.~E.
  (2017).
\newblock Neural message passing for quantum chemistry.
\newblock In {\em International Conference on Machine Learning (ICML)}, pages
  1263--1272. PMLR.

\bibitem[Gondara and Wang, 2018]{gondara2018mida}
Gondara, L. and Wang, K. (2018).
\newblock Mida: Multiple imputation using denoising autoencoders.
\newblock In {\em Pacific-Asia conference on knowledge discovery and data
  mining}, pages 260--272. Springer.

\bibitem[Hamilton et~al., 2017]{hamilton2017inductive}
Hamilton, W., Ying, Z., and Leskovec, J. (2017).
\newblock Inductive representation learning on large graphs.
\newblock {\em Advances in neural information processing systems}, 30.

\bibitem[Jang et~al., 2016]{jang2016categorical}
Jang, E., Gu, S., and Poole, B. (2016).
\newblock Categorical reparameterization with gumbel-softmax.
\newblock {\em arXiv preprint arXiv:1611.01144}.

\bibitem[Jiang and Zhang, 2020]{jiang2020incomplete}
Jiang, B. and Zhang, Z. (2020).
\newblock Incomplete graph representation and learning via partial graph neural
  networks.
\newblock {\em arXiv preprint arXiv:2003.10130}.

\bibitem[Kazi et~al., 2022]{kazi2022differentiable}
Kazi, A., Cosmo, L., Ahmadi, S.-A., Navab, N., and Bronstein, M. (2022).
\newblock Differentiable graph module (dgm) for graph convolutional networks.
\newblock {\em IEEE Transactions on Pattern Analysis and Machine Intelligence}.

\bibitem[Kipf and Welling, 2017]{kipf2016semi}
Kipf, T.~N. and Welling, M. (2017).
\newblock Semi-supervised classification with graph convolutional networks.
\newblock In {\em International Conference on Learning Representations (ICLR)}.

\bibitem[Kool et~al., 2019]{kool2019stochastic}
Kool, W., Van~Hoof, H., and Welling, M. (2019).
\newblock Stochastic beams and where to find them: The gumbel-top-k trick for
  sampling sequences without replacement.
\newblock In {\em International Conference on Machine Learning}, pages
  3499--3508. PMLR.

\bibitem[Kreindler and Lumsden, 2006]{kreindler2006effects}
Kreindler, D.~M. and Lumsden, C.~J. (2006).
\newblock The effects of the irregular sample and missing data in time series
  analysis.
\newblock {\em Nonlinear dynamics, psychology, and life sciences}.

\bibitem[Lakshminarayan et~al., 1996]{lakshminarayan1996imputation}
Lakshminarayan, K., Harp, S.~A., Goldman, R.~P., Samad, T., et~al. (1996).
\newblock Imputation of missing data using machine learning techniques.
\newblock In {\em KDD}, volume~96.

\bibitem[Little and Rubin, 2019]{little2019statistical}
Little, R.~J. and Rubin, D.~B. (2019).
\newblock {\em Statistical analysis with missing data}, volume 793.
\newblock John Wiley \& Sons.

\bibitem[Mackinnon, 2010]{mackinnon2010use}
Mackinnon, A. (2010).
\newblock The use and reporting of multiple imputation in medical research--a
  review.
\newblock {\em Journal of internal medicine}, 268(6):586--593.

\bibitem[Miao et~al., 2022]{miao2022experimental}
Miao, X., Wu, Y., Chen, L., Gao, Y., and Yin, J. (2022).
\newblock An experimental survey of missing data imputation algorithms.
\newblock {\em IEEE Transactions on Knowledge and Data Engineering}.

\bibitem[Muzellec et~al., 2020]{muzellec2020missing}
Muzellec, B., Josse, J., Boyer, C., and Cuturi, M. (2020).
\newblock Missing data imputation using optimal transport.
\newblock In {\em International Conference on Machine Learning}, pages
  7130--7140. PMLR.

\bibitem[Narang et~al., 2013]{narang2013signal}
Narang, S.~K., Gadde, A., and Ortega, A. (2013).
\newblock Signal processing techniques for interpolation in graph structured
  data.
\newblock In {\em 2013 IEEE International Conference on Acoustics, Speech and
  Signal Processing}, pages 5445--5449. IEEE.

\bibitem[Nazabal et~al., 2020]{nazabal2020handling}
Nazabal, A., Olmos, P.~M., Ghahramani, Z., and Valera, I. (2020).
\newblock Handling incomplete heterogeneous data using vaes.
\newblock {\em Pattern Recognition}, 107:107501.

\bibitem[Rossi et~al., 2021]{rossi2021unreasonable}
Rossi, E., Kenlay, H., Gorinova, M.~I., Chamberlain, B.~P., Dong, X., and
  Bronstein, M. (2021).
\newblock On the unreasonable effectiveness of feature propagation in learning
  on graphs with missing node features.
\newblock {\em arXiv preprint arXiv:2111.12128}.

\bibitem[Rubin, 1976]{rubin1976inference}
Rubin, D.~B. (1976).
\newblock Inference and missing data.
\newblock {\em Biometrika}, 63(3):581--592.

\bibitem[Schroff et~al., 2015]{schroff2015facenet}
Schroff, F., Kalenichenko, D., and Philbin, J. (2015).
\newblock Facenet: A unified embedding for face recognition and clustering.
\newblock In {\em Proceedings of the IEEE conference on computer vision and
  pattern recognition}, pages 815--823.

\bibitem[{\'S}mieja et~al., 2018]{smieja2018processing}
{\'S}mieja, M., Struski, {\L}., Tabor, J., Zieli{\'n}ski, B., and Spurek, P.
  (2018).
\newblock Processing of missing data by neural networks.
\newblock {\em Advances in neural information processing systems}, 31.

\bibitem[Spinelli et~al., 2020]{spinelli2020missing}
Spinelli, I., Scardapane, S., and Uncini, A. (2020).
\newblock Missing data imputation with adversarially-trained graph
  convolutional networks.
\newblock {\em Neural Networks}, 129:249--260.

\bibitem[Stekhoven and B{\"u}hlmann, 2012]{stekhoven2012missforest}
Stekhoven, D.~J. and B{\"u}hlmann, P. (2012).
\newblock Missforest—non-parametric missing value imputation for mixed-type
  data.
\newblock {\em Bioinformatics}, 28(1):112--118.

\bibitem[Sterne et~al., 2009]{sterne2009multiple}
Sterne, J.~A., White, I.~R., Carlin, J.~B., Spratt, M., Royston, P., Kenward,
  M.~G., Wood, A.~M., and Carpenter, J.~R. (2009).
\newblock Multiple imputation for missing data in epidemiological and clinical
  research: potential and pitfalls.
\newblock {\em Bmj}, 338.

\bibitem[Taguchi et~al., 2021]{taguchi2021graph}
Taguchi, H., Liu, X., and Murata, T. (2021).
\newblock Graph convolutional networks for graphs containing missing features.
\newblock {\em Future Generation Computer Systems}, 117:155--168.

\bibitem[Troyanskaya et~al., 2001]{troyanskaya2001missing}
Troyanskaya, O., Cantor, M., Sherlock, G., Brown, P., Hastie, T., Tibshirani,
  R., Botstein, D., and Altman, R.~B. (2001).
\newblock Missing value estimation methods for dna microarrays.
\newblock {\em Bioinformatics}, 17(6):520--525.

\bibitem[Van~Buuren, 2018]{van2018flexible}
Van~Buuren, S. (2018).
\newblock {\em Flexible imputation of missing data}.
\newblock CRC press.

\bibitem[Van~Buuren and Groothuis-Oudshoorn, 2011]{van2011mice}
Van~Buuren, S. and Groothuis-Oudshoorn, K. (2011).
\newblock mice: Multivariate imputation by chained equations in r.
\newblock {\em Journal of statistical software}, 45:1--67.

\bibitem[Veli{\v{c}}kovi{\'c} et~al., 2018]{velivckovic2018graph}
Veli{\v{c}}kovi{\'c}, P., Cucurull, G., Casanova, A., Romero, A., Li{\`o}, P.,
  and Bengio, Y. (2018).
\newblock Graph attention networks.
\newblock In {\em International Conference on Learning Representations}.

\bibitem[Vincent et~al., 2008]{vincent2008extracting}
Vincent, P., Larochelle, H., Bengio, Y., and Manzagol, P.-A. (2008).
\newblock Extracting and composing robust features with denoising autoencoders.
\newblock In {\em Proceedings of the 25th international conference on Machine
  learning}, pages 1096--1103.

\bibitem[Wang et~al., 2006]{wang2006missing}
Wang, X., Li, A., Jiang, Z., and Feng, H. (2006).
\newblock Missing value estimation for dna microarray gene expression data by
  support vector regression imputation and orthogonal coding scheme.
\newblock {\em BMC bioinformatics}, 7(1):1--10.

\bibitem[Wang et~al., 2019]{wang2019dynamic}
Wang, Y., Sun, Y., Liu, Z., Sarma, S.~E., Bronstein, M.~M., and Solomon, J.~M.
  (2019).
\newblock Dynamic graph cnn for learning on point clouds.
\newblock {\em Acm Transactions On Graphics (tog)}, 38(5):1--12.

\bibitem[Wu et~al., 2017]{wu2017sampling}
Wu, C.-Y., Manmatha, R., Smola, A.~J., and Krahenbuhl, P. (2017).
\newblock Sampling matters in deep embedding learning.
\newblock In {\em Proceedings of the IEEE international conference on computer
  vision}, pages 2840--2848.

\bibitem[Wu et~al., 2019]{wu2019simplifying}
Wu, F., Souza, A., Zhang, T., Fifty, C., Yu, T., and Weinberger, K. (2019).
\newblock Simplifying graph convolutional networks.
\newblock In {\em International conference on machine learning}, pages
  6861--6871. PMLR.

\bibitem[Yoon et~al., 2016]{yoon2016discovery}
Yoon, J., Davtyan, C., and van~der Schaar, M. (2016).
\newblock Discovery and clinical decision support for personalized healthcare.
\newblock {\em IEEE journal of biomedical and health informatics},
  21(4):1133--1145.

\bibitem[Yoon et~al., 2018a]{yoon2018gain}
Yoon, J., Jordon, J., and Schaar, M. (2018a).
\newblock Gain: Missing data imputation using generative adversarial nets.
\newblock In {\em International conference on machine learning}, pages
  5689--5698. PMLR.

\bibitem[Yoon et~al., 2018b]{yoon2018personalized}
Yoon, J., Zame, W.~R., Banerjee, A., Cadeiras, M., Alaa, A.~M., and van~der
  Schaar, M. (2018b).
\newblock Personalized survival predictions via trees of predictors: An
  application to cardiac transplantation.
\newblock {\em PloS one}, 13(3):e0194985.

\bibitem[Zou et~al., 2019]{zou2019layer}
Zou, D., Hu, Z., Wang, Y., Jiang, S., Sun, Y., and Gu, Q. (2019).
\newblock Layer-dependent importance sampling for training deep and large graph
  convolutional networks.
\newblock {\em Advances in neural information processing systems}, 32.

\end{thebibliography}

\end{document}